\documentclass[10pt,twocolumn,letterpaper]{article}

\usepackage[pagenumbers]{wacv} 

\usepackage[accsupp]{axessibility}  
\usepackage{graphicx}
\usepackage{booktabs}
\usepackage{makecell}
\usepackage{multirow}
\usepackage{pifont}
\usepackage{algorithm}
\usepackage{algpseudocode}

\newcommand{\cmark}{\ding{51}} 
\newcommand{\xmark}{\ding{55}} 

\usepackage{xcolor}

\definecolor{wacvblue}{rgb}{0.21,0.49,0.74}
\usepackage[pagebackref,breaklinks,colorlinks,allcolors=wacvblue,hypertexnames=false]{hyperref}

\def\wacvPaperID{3505} 
\def\confName{WACV}
\def\confYear{2027}

\title{DenseTwins: Dense Modality Decoupling for Multimodal Self-Supervision}

\author{
Carlo Sgaravatti\textsuperscript{1,*} \qquad
Daniele Kopyshevskiy\textsuperscript{1,2,*} \qquad
Davide Secco\textsuperscript{3} \qquad
Giacomo Boracchi\textsuperscript{1}\\
\textsuperscript{1}Politecnico di Milano \qquad
\textsuperscript{2}Sorbonne University \qquad
\textsuperscript{3}Università Bocconi
}

\begin{document}
\maketitle
{\def\thefootnote{}\footnotetext{\raggedright\scriptsize\textsuperscript{*}Equal contribution. Corresponding: \texttt{carlo.sgaravatti@polimi.it}, \texttt{daniele.kopyshevskiy@sorbonne-university.fr}, \texttt{davide.secco@unibocconi.it}}}

\begin{abstract}
Multimodal perception is central to many computer vision applications, including robotics and autonomous driving, where different sensing modalities provide complementary cues.
Self-supervised learning offers a scalable solution for exploiting large amounts of unlabeled multimodal data. However, existing methods either neglect modality-specific information by enforcing feature-space alignment between modalities or learn only global representations without preserving spatial structure.
We propose DenseTwins, a multimodal self-supervised framework that learns dense representations by explicitly decoupling the dense feature space into shared and modality-specific components. The proposed approach combines dense cross-modal correspondences with redundancy-reduction objectives, enabling the model to capture both shared semantics and modality-specific cues within a structured dense embedding space. DenseTwins jointly optimizes an intra-modal objective across augmented views and a cross-modal objective that aligns shared features while decorrelating modality-specific information.
We evaluate DenseTwins across multiple multimodal benchmarks and sensing setups, achieving state-of-the-art or competitive performance across multimodal self-supervised baselines. Code available at: \url{https://github.com/CarloSgaravatti/DenseTwins}.
\end{abstract}
    
\section{Introduction}
\label{sec:intro}

\begin{figure*}[t]
    \centering
    \includegraphics[width=0.8\textwidth]{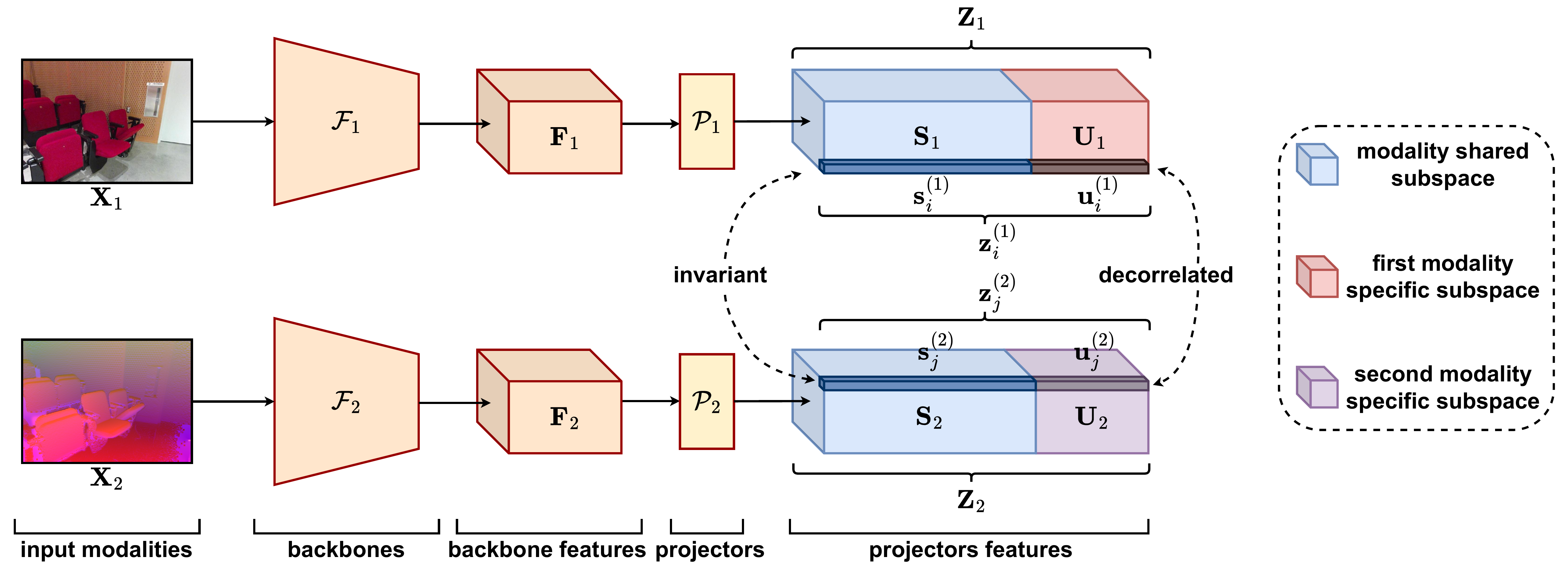}
    \caption{The structure of the dense embeddings of DenseTwins. Two separate encoders $\mathcal{F}_1$, $\mathcal{F}_2$ process the two input modalities $\mathbf{X}_1, \mathbf{X}_2$ separately, producing two feature maps $\mathbf{F}_1, \mathbf{F}_2$. Finally, two projection heads project $\mathbf{F}_1, \mathbf{F}_2$ into a common latent space $\mathbf{Z}_1, \mathbf{Z}_2$, each divided into modality-shared $\mathbf{S}_1, \mathbf{S}_2$ and modality-specific $\mathbf{U}_1, \mathbf{U}_2$ components.}
    \label{fig:embeddings}
\end{figure*}

Many real-world perception systems rely on multiple sensing modalities, such as RGB cameras, depth sensors, LiDAR, event cameras, or thermal imagery \cite{bayoudh2024survey}.
These modalities provide complementary information about the same scene, and their joint exploitation is crucial in perception applications such as autonomous driving, robotics, and video understanding \cite{8269806,sgaravatti2026lcf3d}.
Unfortunately, collecting large-scale annotated multimodal datasets is expensive and often impractical. Self-supervised learning (SSL) provides a natural alternative, enabling learning latent representations directly from unlabeled data \cite{zong2024self}.

SSL has been widely studied in single-modal settings \cite{chen2020simclr,zbontar2021barlow,bardes2021vicreg}. A common approach to extend SSL to multiple modalities is to project inputs from different modalities into a shared embedding space and align their representations. The most prominent example is CLIP~\cite{radford2021learning}, which demonstrates that contrastive learning can effectively align images and text in the latent space. Alignment can also be achieved through redundancy-reduction objectives \cite{zbontar2021barlow,bardes2021vicreg}, which avoid representation collapse by decorrelating feature dimensions. However, in multimodal perception, different sensors provide only partially overlapping information: while some semantic content is shared across modalities, other cues are inherently modality-specific (e.g., texture in RGB, discontinuities in depth images, thermal signatures, or event dynamics). Therefore, an effective multimodal representation should preserve both the shared and the modality-specific information \cite{wang2024decur}.

This motivates \emph{modality decoupling}, where the embedding space is explicitly separated into shared and modality-specific components. To the best of our knowledge, DeCUR~\cite{wang2024decur} is the only multimodal SSL method that explicitly performs decoupling. However, it operates on global representations of the input modalities, discarding spatial structure and limiting the applicability to downstream tasks such as semantic segmentation and object detection, which are particularly relevant in multimodal perception. On the other hand, \emph{dense} approaches that learn pixel-level representations have been studied mainly in the single-modal setting, e.g., DenseCL \cite{wang2021densecl}.

In this work, we propose \textbf{DenseTwins} (\Cref{fig:embeddings}), the first SSL framework explicitly performing decoupling in a dense latent space, preserving the spatial structure while retaining both shared and modality-specific information.
Two modality-specific encoders extract dense feature maps from the input modalities, which are then projected into a structured embedding space.
Part of the embedding space captures information shared across modalities (the \emph{shared subspace}), while the remaining dimensions preserve modality-specific cues (the \emph{modality-specific subspace}). This decomposition allows the model to learn spatially consistent multimodal features while retaining complementary information from each sensing modality. 
Our experiments demonstrate that resulting representations are particularly well-suited for dense perception tasks, while remaining general-purpose representations transferable to global downstream tasks.

DenseTwins builds on three established principles: \emph{redundancy reduction} from Barlow Twins~\cite{zbontar2021barlow}, hence its name, \emph{modality decoupling}~\cite{wang2024decur}, and \emph{dense correspondence learning}~\cite{wang2021densecl}, combining them within a unified multimodal SSL framework.
Distinctive design choices that are key for DenseTwins include: (i) designing a structured dense embedding space that separates shared and modality-specific components spatially, (ii) establishing patch-level cross-modal correspondences within the shared subspace, and (iii) combining dense intra-modal and cross-modal training losses coherently.
Specifically, we first establish dense correspondences across modalities to identify semantically related spatial locations. We then align the shared information while decorrelating the modality-specific content. This design enables the model to learn spatial multimodal representations that capture both shared semantics and modality-specific cues. Since correspondences are estimated directly in feature space, DenseTwins only requires paired observations of the same scene, and can be applied to arbitrary spatially organized modality pairs without modality-specific design choices.

\smallskip\noindent\textbf{Contributions.}
Our main contributions are as follows:
\begin{itemize}
    \item We propose DenseTwins, a multimodal self-supervised framework applicable across heterogeneous spatially organized modalities that jointly learns \textbf{dense representations} that \textbf{explicitly decouple} different modalities.
    \item We introduce a cross-modal training objective aligning the shared components of semantically corresponding features while preserving modality-specific information in the embeddings.
\end{itemize}
We evaluate DenseTwins across multiple multimodal datasets and sensing setups (RGB--DEM, RGB--Depth, RGB--LiDAR, RGB--Event, SAR--Optical, RGB--Thermal). Experimental results show that DenseTwins consistently improves overall performance across 2D and 3D semantic segmentation, object detection, and multilabel classification downstream tasks.
\section{Related Work}
\label{sec:related-work}

SSL methods can be broadly categorized into two groups: (i) \emph{Global SSL} methods, which learn global representations of the inputs, and (ii) \emph{Dense SSL} methods, which learn representations at the pixel or patch level. Orthogonally, in multimodal learning, \emph{Modality Decoupling} approaches explicitly separate the information shared across modalities from the information that is specific to each modality.

\subsection{Global self-supervised representation learning.}
Among global SSL methods, contrastive approaches such as SimCLR~\cite{chen2020simclr} maximize agreement between pairs of augmented views of the same input (positive pairs) while pushing apart the negative samples given by the other samples in the mini-batch (negatives). 
MoCo~\cite{he2020moco} introduced a momentum encoder and a queue-based dictionary that allows defining negative samples also outside the mini-batch to scale contrastive learning without increasing the batch size. MoCoV2~\cite{he2020mocov2} adopts stronger augmentations and custom projection heads applied to the global representations. 
Redundancy reduction approaches such as Barlow Twins~\cite{zbontar2021barlow} and VICReg~\cite{bardes2021vicreg} learn representations by enforcing invariance of the feature space to different augmentations.
While these frameworks are designed for processing single-modal inputs, they can be extended to multimodal settings. Similarly,
CLIP~\cite{radford2021learning} demonstrated the effectiveness of large-scale cross-modal contrastive pretraining, aligning global representations across modalities. However, all these alignment strategies focus on modality-shared features and may suppress useful modality-specific information. Moreover, they do not explicitly model spatial structure, limiting their effectiveness for downstream dense prediction tasks.

\subsection{Dense self-supervised representation learning.}
DenseCL~\cite{wang2021densecl} introduced dense contrastive learning by establishing correspondences between spatial locations across different augmented views of the same input image, by maximizing their cosine similarity in the feature space.
DenseCL++~\cite{iskender2022improving} improves correspondences and training stability, further enhancing performance on dense downstream tasks. Differently, PixelPro \cite{xie2021propagate} computes the normalized distances between features of the pixels of two augmented views and defines positive and negative pairs using a predefined threshold.
Recent self-distillation methods such as DINO~\cite{caron2021emerging}, iBOT~\cite{zhou2021ibot}, and DINOv2~\cite{oquab2023dinov2} learn representations by matching the outputs of a teacher and a student network under different image augmentations, leveraging either a local-to-global view matching objective \cite{caron2021emerging,oquab2023dinov2}, or a masked online tokenization objective \cite{zhou2021ibot,oquab2023dinov2}, yielding spatially structured dense representations with remarkable emergent semantic properties. Predictive learning frameworks such as I-JEPA~\cite{assran2023self} and LeJEPA~\cite{balestriero2025lejepa} instead learn dense representations by predicting latent representations of masked image regions, with LeJEPA further providing a provable, heuristic-free training objective.
These approaches are designed for single-modal settings, and their extension to arbitrary multimodal input pairs is non-trivial. In contrast, DenseTwins establishes dense correspondences across modalities and applies modality decoupling at the patch level, jointly matching and separating shared and modality-specific content within dense representations.

\subsection{Modality decoupling.}
Modality decoupling in SSL has been studied by DeCUR~\cite{wang2024decur}, which explicitly decouples common and modality-specific components using a redundancy-reduction objective analogous to Barlow Twins. However, DeCUR~\cite{wang2024decur} operates on global feature vectors and does not preserve spatial structure, limiting its applicability to dense prediction tasks. Other approaches apply modality decoupling for sentiment analysis and person re-identification ~\cite{Hazarika2020MISAMA,Lai2023SharedAP,hao2021cross}, as well as across audio, video and text~\cite{alayrac2020self,wang2024information}; however, these methods are either supervised or designed for non-visual, non-dense tasks.
\section{Problem Formulation}
\label{sec:problem-formulation}

We consider a multimodal self-supervised learning setting, where each training sample consists of observations of the same scene acquired through different sensing modalities. 
Formally, we are given an unlabeled dataset
\[
\mathcal{D} = \{(\mathbf{X}_1^{(i)}, \mathbf{X}_2^{(i)})\}_{i=1}^{N},
\]
where $\mathbf{X}_m^{(i)}$ denotes the observation of the $i$-th scene captured by modality $m \in \{1, 2\}$. 
The two modalities may correspond to heterogeneous sensors (e.g., RGB, depth, thermal, LiDAR, or event-camera acquisitions) and are not required to be perfectly spatially aligned, even though we require synchronized acquisition time.

Our goal, as illustrated in \Cref{fig:embeddings}, is to learn modality-specific encoders $\mathcal{F}_{1}, \mathcal{F}_{2}$, where each encoder $\mathcal{F}_m : \mathbf{X}_m \rightarrow \mathbf{F}_m$ maps an input observation $\mathbf{X}_m$ to a dense latent representation $\mathbf{F}_m$ containing $T_m$ feature vectors $\mathbf{f}_{m,j} \in \mathbb{R}^{K}$.
We consider any encoder $\mathcal{F}_{m}$ producing spatially organized representations; thus each vector $\mathbf{f}_{m,j} \in \mathbb{R}^{K}$ corresponds to a localized input region, e.g., a CNN feature-map element (ResNet) or a token of a vision Transformer (ViT). Non-spatial modalities, such as text, are therefore outside the scope of our formulation.
Our primary objective is to train $\mathcal{F}_{1}$ and $\mathcal{F}_{2}$ to learn representations suitable for addressing downstream \emph{dense prediction tasks}, such as semantic segmentation.
The challenge is to learn such representations in a fully self-supervised manner, without relying on manual annotations, and from different sensing modalities that often capture complementary information.
\section{DenseTwins}
\label{sec:method}

\begin{figure*}[t]
\centering
\includegraphics[width=0.9\textwidth]{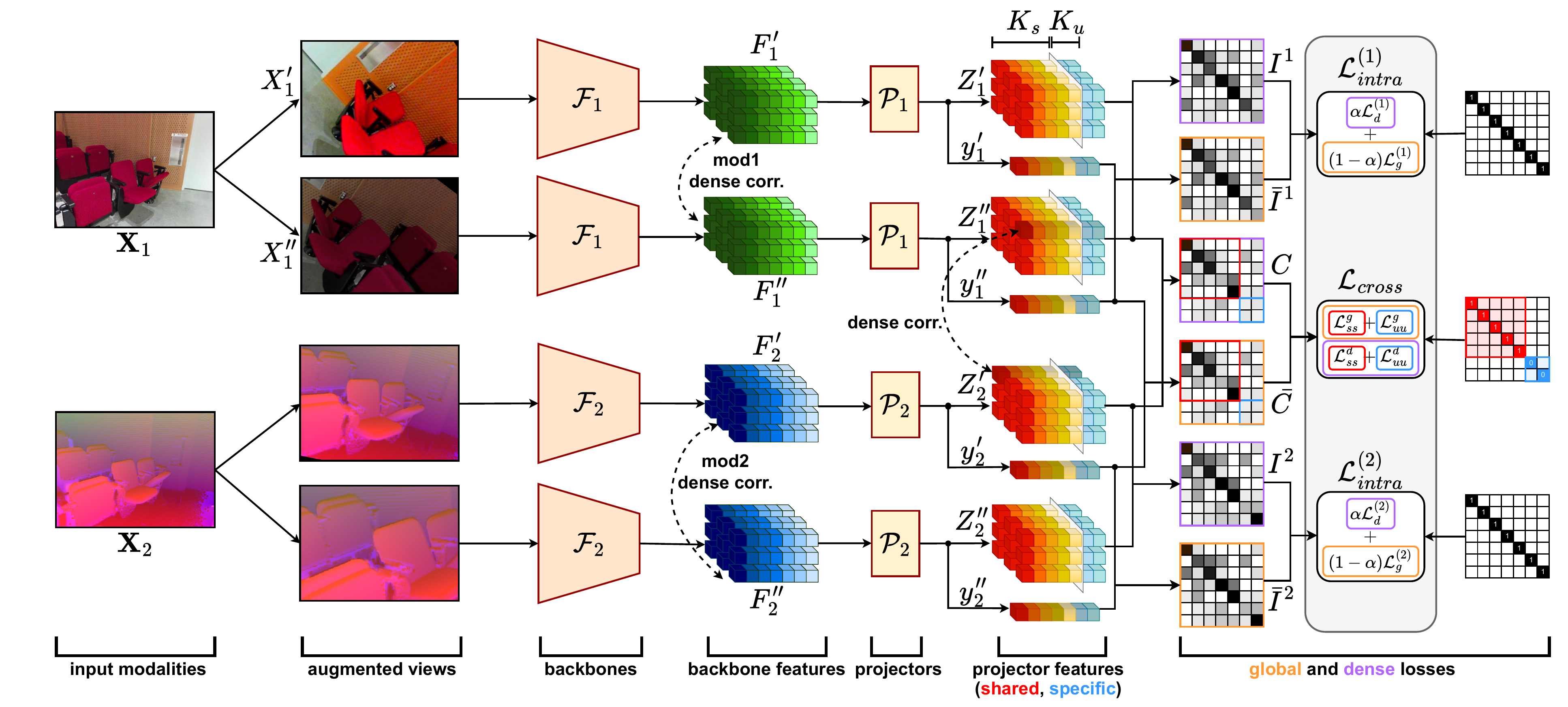}
\caption{\textbf{DenseTwins architecture}. 
For each modality $m \in \{1, 2\}$, two augmented views are processed by a modality-specific backbone $\mathcal{F}_m$, followed by a projection head $\mathcal{P}_m$ that produces both global and dense embeddings. 
The embedding space is explicitly partitioned into a shared subspace (red) and a modality-specific subspace (blue). 
Training is based on dense correspondences between feature vectors within each modality and across modalities computed using cosine similarities. 
The shared components are aligned through redundancy reduction and invariance objectives, while modality-specific components are decorrelated across modalities to preserve complementary information. Intra-modal learning, instead, applies a redundancy-reduction objective, encouraging invariance between augmented views.
The final objective combines intra-modal and cross-modal losses applied to both global and dense representations.}
\label{fig:architecture}
\end{figure*}

\Cref{fig:architecture} illustrates the DenseTwins architecture, which consists of two separate branches, one per modality.
For each modality input $\mathbf{X}_m$, two independently augmented views $(\mathbf{X}_m^{'}, \mathbf{X}_m^{''})$ of $\mathbf{X}_m$ are processed by the same encoder $\mathcal{F}_m$, extracting feature maps $(\mathbf{F}_m^{'}, \mathbf{F}_m^{''})$. These are fed to a projection layer $\mathcal{P}_m$ that is composed of: (i) a \emph{global projection head} $\mathcal{P}_m^g$ and (ii) a \emph{dense projection head} $\mathcal{P}_m^d$, both projecting into a $K_p$-dimensional feature space. The \emph{global projection head} extracts the global embeddings from $(\mathbf{F}_m^{'}, \mathbf{F}_m^{''})$ by global average pooling, and projects them to $K_p$-dimensional vectors $\mathbf{y}_m^{'}, \mathbf{y}_m^{''}$. The \emph{dense projection head} preserves spatial structure, necessary for downstream dense prediction tasks, by computing the dense feature maps $(\mathbf{Z}_m^{'}, \mathbf{Z}_m^{''})$. Following common practice in SSL, the projection heads operate in a high-dimensional embedding space. This design has been shown to improve representation learning, particularly for redundancy-reduction objectives such as Barlow Twins~\cite{zbontar2021barlow}.

Inspired by DeCUR \cite{wang2024decur}, we partition the embedding space into a \emph{shared} subspace (first $K_s$ components) capturing cross-modal semantics, and a \emph{modality-specific} subspace (last $K_u$ components) preserving unique cues (\Cref{fig:architecture}), extending this decomposition, applied only to global representations in DeCUR, to dense representations. We establish spatial correspondences across modalities by measuring similarity in the shared subspace, while enforcing decorrelation of the modality-specific components.

\smallskip\noindent\textbf{Downstream inference}. Only $\mathcal{F}_1,\mathcal{F}_2$ are transferred to downstream tasks: $\mathcal{P}_1,\mathcal{P}_2$ are discarded after pretraining. The separation between $\mathcal{F}_1$ and $\mathcal{F}_2$ makes DenseTwins agnostic to the fusion strategy; either simple feature concatenation or task-specific architectures can be employed.


\smallskip\noindent\textbf{Cross-correlation primitive.}
All training objectives in DenseTwins share a common building block. Given a set of embedding pairs $\{(\mathbf{a}_b, \mathbf{b}_b)\}_{b=1}^{B}$ with $\mathbf{a}_b, \mathbf{b}_b \in \mathbb{R}^{K_p}$, we compute a cross-correlation matrix $\Sigma \in \mathbb{R}^{K_p \times K_p}$:
\begin{equation}\label{eq:correlation}
\Sigma_{pq} = \frac{\sum_{b} a_{b,p}\, b_{b,q}}{\sqrt{\sum_{b} a_{b,p}^2}\;\sqrt{\sum_{b} b_{b,q}^2}},
\end{equation}
DenseTwins uses four such matrices: two intra-modal matrices $\bar{I}^{(m)}$ and $I^{(m)}$, computed from global and dense embeddings of the same modality, and two cross-modal matrices $\bar{C}$ and $C$, computed from embeddings of different modalities. \Cref{tab:matrices} summarises the correlation matrices of DenseTwins, all having size $K_p \times K_p$.
For global matrices ($\bar{I}^{(m)}, \bar{C}$), $B$ equals the batch size; for dense matrices ($I^{(m)}, C$), $B$ equals the batch size multiplied by the number of correspondences.

\begin{table}[t]
\centering
\small
\begin{tabular}{lll}
\toprule
\textbf{Matrix} & \textbf{Input pairs} & \textbf{Section} \\
\midrule
$\bar{I}^{(m)}$ & global intra-modal embeddings     & \cref{sec:intra} \\
$I^{(m)}$       & dense intra-modal correspondences & \cref{sec:intra} \\
$\bar{C}$       & global cross-modal embeddings     & \cref{sec:cross} \\
$C$             & dense cross-modal correspondences & \cref{sec:cross} \\
\bottomrule
\end{tabular}
\caption{Cross-correlation matrices in DenseTwins.}
\label{tab:matrices}
\end{table}

Since each projected embedding is decomposed into a shared component of dimension $K_s$ and a modality-specific component of dimension $K_u$ (with $K_s + K_u = K_p$, illustrated in \Cref{fig:embeddings}), the cross-modal matrices $\bar{C}$ and $C$ admit the following block partition:
\begin{equation}\label{eq:correlation-blocks}
\bar{C} = \begin{bmatrix} \bar{C}^{ss} & \bar{C}^{su} \\ \bar{C}^{us} & \bar{C}^{uu} \end{bmatrix}, \qquad
C = \begin{bmatrix} C^{ss} & C^{su} \\ C^{us} & C^{uu} \end{bmatrix},
\end{equation}
where $(\cdot)^{ss} \in \mathbb{R}^{K_s \times K_s}$ captures correlations within the shared subspace, and $(\cdot)^{uu} \in \mathbb{R}^{K_u \times K_u}$ within the modality-specific subspace.

The overall training objective is computed as follows:
\begin{equation}
\mathcal{L}_{\text{total}} = \mathcal{L}_{\text{intra}}^{(1)} + \mathcal{L}_{\text{intra}}^{(2)} + \mathcal{L}_{\text{cross}},
\end{equation}
where $\mathcal{L}_{\text{intra}}^{(1)}$ and $\mathcal{L}_{\text{intra}}^{(2)}$ assess consistency between augmented views of the same modality (\cref{sec:intra}), and $\mathcal{L}_{\text{cross}}$ aligns the shared subspace while decorrelating the modality-specific ones (\cref{sec:cross}).

\subsection{Intra-Modal Loss}
\label{sec:intra}
As in SimCLR \cite{chen2020simclr}, two augmented views of the same modality are generated, and we replicate this process for each modality. Intra-modal learning enforces consistency between different views of the same image, learning invariant representations that capture semantic content. 

\smallskip\noindent\textbf{Intra-Modal Loss.} To achieve invariance in the augmented views, the intra-modal loss is a convex combination of both global and dense redundancy reduction losses \cite{zbontar2021barlow}:
\begin{equation}
    \mathcal{L}_{\text{intra}}^{(m)} = (1 - \alpha)\mathcal{L}_{g}^{(m)} + \alpha\mathcal{L}_{d}^{(m)},
\end{equation}
where $\alpha \in [0,1]$ balances $\mathcal{L}_g^{(m)}$ and $\mathcal{L}_d^{(m)}$. This loss is computed for each modality $m \in \{1, 2\}$ independently, resulting in two intra-modal losses, $\mathcal{L}_{\text{intra}}^{(1)}$ and $\mathcal{L}_{\text{intra}}^{(2)}$.

\smallskip\noindent\textbf{Global redundancy-reduction.} The global redundancy reduction loss is applied to the global embeddings within the same modality. Given a batch of paired global embeddings $(\mathbf{y}^{'}, \mathbf{y}^{''})$ of the two augmented views from the same modality, we compute $\bar{I}^{(m)}$ as in \eqref{eq:correlation}. The loss is then defined as
\begin{equation}\label{eq:barlow-twins}
\mathcal{L}_{g}^{(m)} =
\sum_{p} (1 - \bar{I}_{pp}^{(m)})^2
+
\lambda \sum_{p \neq q} (\bar{I}_{pq}^{(m)})^2.
\end{equation}
By minimizing $\mathcal{L}_g^{(m)}$, we enforce: (i) \textbf{Invariance}, by pushing diagonal entries toward one to ensure both views have similar embeddings, and (ii) \textbf{Redundancy reduction}, by forcing the off-diagonal entries towards zero.

\smallskip\noindent\textbf{Dense token alignment and redundancy reduction.}
Let $\mathbf{F}'$ and $\mathbf{F}''$ be spatial features returned by the backbone $\mathcal{F}$ from the two augmented views of the same modality:
\[
\mathbf{F}' = \{\mathbf{f}_{i}'\}_{i=1}^{T}, 
\quad 
\mathbf{F}'' = \{\mathbf{f}_{j}''\}_{j=1}^{T},
\quad 
\mathbf{f}_{i}', \mathbf{f}_{j}'' \in \mathbb{R}^{K}.
\]
Similarly, let $\mathbf{Z}'$ and $\mathbf{Z}''$ be the output of the dense projection head for the two augmented views:
\[
\mathbf{Z}' = \{\mathbf{z}_{i}'\}_{i=1}^{T}, 
\quad 
\mathbf{Z}'' = \{\mathbf{z}_{j}''\}_{j=1}^{T},
\quad 
\mathbf{z}_{i}', \mathbf{z}_{j}'' \in \mathbb{R}^{K_p}.
\]
To extend redundancy reduction to dense representation, we establish correspondences between every single spatial feature $\mathbf{z}_{i}' \in \mathbf{Z}'$ and one vector of $\mathbf{Z}''$. This association needs to preserve semantics: we require the two corresponding tokens to refer to semantically similar parts of the input (e.g. regions referring to the same type of object in RGB images). However, augmentations alter the spatial layouts, and tokens at the same spatial index might not correspond to the same semantic region. 

To address this spatial misalignment, we establish correspondences in the backbone representation space by computing the pairwise cosine similarity between embeddings from the two augmented views:
\begin{equation}\label{eq:cosine}
\Delta_{ij} =
\frac{\mathbf{f}_{i}'^\top \mathbf{f}_{j}''}
{\|\mathbf{f}_{i}'\| \, \|\mathbf{f}_{j}''\|}.
\end{equation}
We define a set of matches approximating semantic correspondences across views. In practice, each token $\mathbf{f}_{i}'$ in the first view is matched with the most similar token in the second view:
\begin{equation}\label{eq:correspondence}
j_i^* = \arg\max_{j} \Delta_{ij}.
\end{equation}
We then retrieve the corresponding projected embeddings $(\mathbf{z}_{i}', \mathbf{z}_{j_i^*}'')$ and compute an intra-modality cross-correlation matrix $I^{(m)} \in \mathbb{R}^{K_p \times K_p}$ over matched dense token pairs as in \eqref{eq:correlation}. The redundancy-reduction loss in \eqref{eq:barlow-twins}, which we denote by $\mathcal{L}_{d}^{(m)}$, is applied to $I^{(m)}$.

\subsection{Cross-Modal Loss}
\label{sec:cross}
Cross-modal learning leverages the relationship between representations extracted from different modalities. 
The objective is twofold: 
(i) align information that is shared across modalities, and (ii) prevent modality-specific information from collapsing into the shared representation.

Let $\mathbf{y}_1, \mathbf{y}_2 \in \mathbb{R}^{K_p}$ denote the global projected embeddings of the two modalities, and 
$\mathbf{Z}_1 = \{\mathbf{z}_{1,i}\}_{i=1}^{T_1}$,
$\mathbf{Z}_2 = \{\mathbf{z}_{2,j}\}_{j=1}^{T_2}$ the dense projected embeddings.

\smallskip\noindent\textbf{Embedding space decomposition.}
The projector embedding space is partitioned into a shared subspace of dimension $K_s$ and a modality-specific subspace of dimension $K_u$, with $K_p = K_s + K_u$, as illustrated in \Cref{fig:embeddings,fig:architecture}:

\[
\mathbf{z} =
\begin{bmatrix}
\mathbf{s} \\
\mathbf{u}
\end{bmatrix},
\quad
\mathbf{s} \in \mathbb{R}^{K_s},
\quad
\mathbf{u} \in \mathbb{R}^{K_u}.
\]
This decomposition is applied to both global $\mathbf{y}_1, \mathbf{y}_2$ and dense $\mathbf{Z}_1, \mathbf{Z}_2$ embeddings.

\smallskip\noindent\textbf{Final cross-modal objective.}
The cross-modal loss function combines global and dense components, both enforcing modality decoupling and decorrelation on global and dense embeddings:

\begin{equation}
\mathcal{L}_{\text{cross}} = 
(1 - \alpha) \mathcal{L}_{\text{cross}}^{g}
+
\alpha \mathcal{L}_{\text{cross}}^{d}.
\end{equation}

\smallskip\noindent\textbf{Global cross-modal loss.}
As in DeCUR~\cite{wang2024decur}, the cross-correlation matrix $\bar{C}$ is computed from global embeddings $(\mathbf{y}_1, \mathbf{y}_2)$ via \eqref{eq:correlation}, and partitioned into $\bar{C}^{ss}$ and $\bar{C}^{uu}$ as in \eqref{eq:correlation-blocks}.
The goal is to align the first $K_s$ components of $\mathbf{y}_1, \mathbf{y}_2$ and make the last $K_u$ components uncorrelated to promote learning modality-specific features. Thus, we apply an invariance and redundancy reduction loss as in \eqref{eq:barlow-twins} to $\bar{C}^{ss}$, and enforce full decorrelation on $\bar{C}^{uu}$:
\begin{align}
\mathcal{L}_{ss}^{g} &=
\sum_{p=1}^{K_s} (1 - \bar{C}^{ss}_{pp})^2
+
\lambda \sum_{p \neq q}^{K_s} (\bar{C}^{ss}_{pq})^2,\\
\mathcal{L}_{uu}^{g} &=
\sum_{p=1}^{K_u} (\bar{C}^{uu}_{pp})^2
+
\lambda \sum_{p \neq q}^{K_u} (\bar{C}^{uu}_{pq})^2.
\end{align}
The global cross-modal loss is finally defined as:
\begin{equation}\label{eq:global-cross-modal-loss}
\mathcal{L}_{\text{cross}}^{g} = \mathcal{L}_{ss}^g + \mathcal{L}_{uu}^g.
\end{equation}

\smallskip\noindent\textbf{Dense cross-modal decoupling.}
The goal for cross-modal dense representation decoupling is similar to the global case: training the shared subspace of the embedding to be invariant across modalities and the modality-specific subspace to be uncorrelated.
Similarly to the intra-modal case, we first establish cross-modal correspondences between tokens of the two modalities. However, as the embeddings are split into modality-shared and specific, we use only the shared subspace to define correspondences since this is the subspace where cross-modal alignment is expected. 

For each feature vector $\mathbf{z}_{1,i}$, we extract its common component $\mathbf{s}_{1,i}$ and compute the cosine similarity $\Delta_{ij}^{\text{cross}}$ with common components $\mathbf{s}_{2,j}$ from the second modality as in \eqref{eq:cosine}.
Then we select, for each spatial location $i$ in the first modality, the most similar location $j_i^*$ in the second modality as in \eqref{eq:correspondence}.
Using these matched token pairs $(\mathbf{z}_{1,i}, \mathbf{z}_{2,j_i^*})$, we compute a cross-correlation matrix $C \in \mathbb{R}^{K_p \times K_p}$ over the full projected embeddings (both shared and modality-specific components), exactly as in the global case, and partition it into blocks $C^{ss}$ and $C^{uu}$, as in \eqref{eq:correlation-blocks}.

The same combination of invariance, redundancy reduction and decorrelation used at the global level in \eqref{eq:global-cross-modal-loss} is then applied to $C^{ss}$ and $C^{uu}$, yielding the following dense cross-modal loss:
\begin{equation}
\mathcal{L}_{\text{cross}}^{d}
=
\mathcal{L}_{ss}^{d}
+
\mathcal{L}_{uu}^{d}.
\end{equation}
As correspondences are noisy early in training, we apply a linear warmup of $N$ epochs to the weight $\alpha$ of $\mathcal{L}_{\text{cross}}^{d}$.
\section{Experiments}
\label{sec:experiments}

\smallskip\noindent\textbf{Datasets.}
We evaluate DenseTwins across six multimodal sensing pairs, covering semantic segmentation, object detection, and multilabel classification: RGB--Depth (SUN-RGBD~\cite{song2015sun}), RGB--Event (EventScape~\cite{RAL21Gehrig}, DSEC~\cite{gehrig2021dsec}), RGB--LiDAR (Semantic-KITTI~\cite{behley2019iccv}), RGB--DEM (GeoNRW~\cite{geonrw}), SAR--Optical (BigEarthNet~\cite{sumbul2019bigearthnet}), and RGB--Thermal (KAIST~\cite{hwang2015kaist}). Preprocessing and pretraining details for each dataset are provided in the supplementary material (\Cref{sec:supp-pretraining}).

\smallskip\noindent\textbf{Competitors.}
We adapt SimCLR~\cite{chen2020simclr}, Barlow Twins~\cite{zbontar2021barlow} and VICReg~\cite{bardes2021vicreg} to the multimodal setting, applying the two modalities in place of the two augmented views normally required by these methods. We also consider CLIP~\cite{radford2021learning} and DeCUR~\cite{wang2024decur}, which are natively multimodal. All these methods operate on global representations; thus, we take the last dense feature map before pooling. Finally, we compare with single-modal methods: DINO~\cite{caron2021emerging}, iBOT~\cite{zhou2021ibot}, I-JEPA~\cite{assran2023self} and LeJEPA~\cite{balestriero2025lejepa}, which we train independently on each modality. Further details are provided in the supplementary material (\Cref{sec:supp-competitors}).

\subsection{Evaluation Protocols}
\label{sec:eval-protocols}

We test under two complementary evaluation protocols (additional details are available in \Cref{sec:supp-finetuning}).

\smallskip\noindent\textbf{Frozen-backbone probing.} We assess the quality of the learned representations by freezing both pretrained encoders and training only a downstream predictor. Since all methods learn one encoder per modality, their feature maps are concatenated before the task-specific head. For semantic segmentation, we evaluate increasingly expressive predictors: k-nearest-neighbours (kNN), linear probing (LP), and a lightweight FCN. kNN and LP operate directly at the feature-map resolution, with predictions subsequently upsampled to the input resolution. For object detection, we train a lightweight FCOS~\cite{tian2019fcos} head on the frozen features, while multilabel classification is evaluated through linear probing. All competing methods use the same backbone to ensure a controlled comparison: unless otherwise specified, we use ViT-S as iBOT and I-JEPA are mainly based on Transformer-based architectures.

\smallskip\noindent\textbf{Fine-tuning.} Following DeCUR~\cite{wang2024decur}, we also evaluate how effectively SSL provides an initialization for downstream training when all transferred parameters are optimized jointly for the downstream task. As in DeCUR, for 2D semantic segmentation, we adopt CMX~\cite{zhang2023cmx} with MiT-B2~\cite{xie2021segformer} as backbone, and FCN~\cite{long2015fully} with ResNet50~\cite{he2016resnet}. For SemanticKITTI, we fine-tune PMF~\cite{zhuang2021perception}, whose RGB branch uses ResNet34 and transfer only the pretrained RGB backbone, as PMF's early RGB-LiDAR fusion strategy is not compatible with an independent LiDAR backbone.

\smallskip\noindent\textbf{Metrics.} We report the mean Intersection over Union (mIoU) for semantic segmentation, Average Precision at 50\% of IoU (AP50) for object detection, and mean Average Precision (mAP) for multilabel classification.

\subsection{2D Semantic Segmentation}

\begin{table}[t]
\centering
\small
\setlength{\tabcolsep}{4pt}
\caption{SUN-RGBD and EventScape (CMX fine-tuning).}
\label{tab:cmx-finetune-results}
\begin{tabular}{lcc c}
\toprule
& \multicolumn{2}{c}{SUN-RGBD mIoU} & EventScape mIoU \\
\cmidrule(lr){2-3}\cmidrule(lr){4-4}
Pretrain & 10\% labels & 100\% labels & 100\% labels \\
\midrule
None & 35.7 & 49.7 & 61.9 \\
\midrule
SimCLR & 37.6 & 50.8 & 63.2 \\
CLIP & 37.6 & 50.6 & \textbf{63.5} \\
BarlowTwins & \underline{37.8} & \underline{51.0} & 63.0 \\
VICReg & 37.7 & \underline{51.0} & 62.4 \\
DeCUR & 36.9 & 50.6 & 62.3 \\
DenseTwins & \textbf{38.2} & \textbf{51.5} & \underline{63.3} \\
\bottomrule
\end{tabular}
\end{table}

We evaluate 2D semantic segmentation on SUN-RGBD~\cite{song2015sun} and EventScape~\cite{RAL21Gehrig}. On SUN-RGBD, we consider both full supervision and a 10\% low-label setting, together with frozen-backbone probing on the reduced 13-class label set~\cite{couprie2013indoor}. EventScape is evaluated under full-supervision fine-tuning.

DenseTwins achieves the best fine-tuned mIoU on SUN-RGBD at both label budgets (\Cref{tab:cmx-finetune-results}). Despite also decoupling modalities, DeCUR underperforms most baselines, suggesting that global-only decoupling is less effective for dense tasks. On EventScape, DenseTwins reaches 63.3 mIoU, above all redundancy-reduction baselines; only CLIP (63.5) scores slightly higher. Qualitative CMX segmentations are in \Cref{sec:supp-qualitative}. Under frozen-backbone probing on SUN-RGBD (\Cref{tab:sunrgbd-probe}), DenseTwins outperforms every baseline by a wide margin, demonstrating the quality of the learned representations independently of downstream fine-tuning. We further analyze single-modal encoders in isolation in \Cref{sec:supp-missing-modalities}, showing competitive results even when only one modality is available at test time.

\begin{table}[t]
\centering
\small
\setlength{\tabcolsep}{4pt}
\caption{SUN-RGBD frozen probing (NYU-13 class label set).}
\label{tab:sunrgbd-probe}
\resizebox{0.6\linewidth}{!}{%
\begin{tabular}{lccc}
\toprule
Pretrain & kNN & LP & FCN \\
\midrule
SimCLR & 22.4 & 25.1 & 30.4 \\
CLIP & 20.3 & 22.3 & 29.9 \\
Barlow Twins & 18.5 & 20.2 & 27.0 \\
VICReg & 18.6 & 22.7 & 28.0 \\
DeCUR & 22.0 & 23.7 & 35.8 \\
\midrule
DINO & 25.6 & 31.0 & \underline{37.6} \\
iBOT & 24.9 & 30.8 & 36.3 \\
I-JEPA & \underline{27.4} & \underline{31.1} & 35.2 \\
LeJEPA & 18.8 & 23.9 & 32.7 \\
\midrule
DenseTwins & \textbf{29.1} & \textbf{33.7} & \textbf{40.6} \\
\bottomrule
\end{tabular}
}
\end{table}

\subsection{Object Detection for Autonomous Driving}

Frozen-backbones evaluation for object detection on KAIST~\cite{hwang2015kaist} and DSEC~\cite{gehrig2021dsec} shows that DenseTwins achieves the best overall AP50 on both benchmarks under the same backbone (\Cref{tab:detection-results}). We additionally report the results of DenseTwins using Swin-T and ConvNeXtV2-T to demonstrate further improvements. These results highlight that the benefits of dense modality decoupling extend beyond semantic segmentation and generalize across backbone families. Notably, global-representation methods substantially lag behind DenseTwins and the dense single-modal methods on both benchmarks, suggesting that object detection benefits more from dense, patch-level pretraining than from a single global representation.

\begin{table}[t]
\centering
\small
\caption{Object detection AP50 on KAIST and DSEC. Baselines use ViT-S/16; DenseTwins also uses Swin-T and ConvNeXtV2-T.}
\label{tab:detection-results}
\setlength{\tabcolsep}{4pt}
\resizebox{\linewidth}{!}{%
\begin{tabular}{llcc ccc}
\toprule
& & \multicolumn{2}{c}{KAIST AP50} & \multicolumn{3}{c}{DSEC AP50} \\
\cmidrule(lr){3-4}\cmidrule(lr){5-7}
Pretrain & Backbone & All & Reasonable & All & Car & Pedestrian \\
\midrule
DenseTwins & Swin-T & \textbf{33.3} & \textbf{35.3} & \textbf{24.3} & \textbf{51.1} & \underline{11.9} \\
DenseTwins & ConvNeXtV2-T & \underline{31.2} & \underline{32.9} & \underline{21.3} & \underline{47.1} & 9.7 \\
DenseTwins & ViT-S/16 & 26.8 & 29.8 & 20.6 & 46.0 & 9.3 \\
\midrule
SimCLR & ViT-S/16 & 16.8 & 19.2 & 8.4 & 36.7 & 6.8 \\
CLIP & ViT-S/16 & 16.5 & 18.8 & 8.2 & 31.0 & 5.1 \\
Barlow Twins & ViT-S/16 & 12.0 & 14.8 & 7.8 & 30.0 & 4.9 \\
VICReg & ViT-S/16 & 13.1 & 15.1 & 8.5 & 34.6 & 5.2 \\
DeCUR & ViT-S/16 & 12.0 & 13.8 & 7.9 & 31.6 & 4.7 \\
DINO & ViT-S/16 & 25.8 & 30.6 & 18.4 & 44.6 & \textbf{12.0} \\
iBOT & ViT-S/16 & 24.8 & 29.2 & 16.2 & 43.5 & 9.2 \\
I-JEPA & ViT-S/16 & 25.8 & 29.2 & 14.9 & 37.9 & 7.8 \\
LeJEPA & ViT-S/16 & 19.8 & 22.4 & 14.1 & 38.1 & 7.0 \\
\bottomrule
\end{tabular}
}
\end{table}

\subsection{Remote Sensing}

We evaluate DenseTwins on GeoNRW~\cite{geonrw} (RGB--DEM segmentation), under both fine-tuning and frozen-probing protocols, and BigEarthNet~\cite{sumbul2019bigearthnet} (SAR--Optical multilabel classification). As shown in \Cref{tab:remote-sensing}, DenseTwins gives the best low-label GeoNRW mIoU and the best frozen-probing results across all three protocols, while remaining competitive with all baselines at full supervision. On BigEarthNet, DenseTwins achieves the best classification performance, outperforming CLIP, SimCLR and DeCUR, confirming that the benefits extend to image-level recognition as well.

\begin{table}[t]
\centering
\small
\caption{GeoNRW (fine-tuned FCN and frozen probing) and BigEarthNet (frozen-probing mAP).}
\label{tab:remote-sensing}
\setlength{\tabcolsep}{4pt}
\resizebox{\linewidth}{!}{%
\begin{tabular}{l cc ccc c}
\toprule
& \multicolumn{2}{c}{GeoNRW (fine-tuned FCN)} & \multicolumn{3}{c}{GeoNRW (frozen probe)} & BigEarthNet \\
\cmidrule(lr){2-3}\cmidrule(lr){4-6}\cmidrule(lr){7-7}
Pretrain & 1\% mIoU & 100\% mIoU & kNN & LP & FCN & mAP \\
\midrule
Random & 26.2 & 35.4 & 16.1 & 19.5 & 30.1 & 0.601 \\
\midrule
SimCLR & 40.6 & 55.8 & 22.5 & 27.9 & 42.9 & 0.809 \\
CLIP & 40.8 & 55.2 & 23.8 & 28.9 & 42.6 & 0.819 \\
BarlowTwins & 40.8 & 52.1 & 19.2 & 23.7 & 41.8 & 0.797 \\
VICReg & 41.0 & \textbf{56.3} & 24.5 & 27.9 & 43.9 & 0.803 \\
DeCUR & \underline{41.5} & 54.8 & \underline{30.6} & \underline{34.7} & \underline{45.5} & \underline{0.831} \\
DenseTwins & \textbf{41.7} & \underline{56.0} & \textbf{36.6} & \textbf{40.5} & \textbf{46.1} & \textbf{0.834} \\
\bottomrule
\end{tabular}
}
\end{table}

\subsection{RGB-LiDAR 3D Semantic Segmentation}

\begin{table}[t]
\centering
\small
\caption{SemanticKITTI: PMF (fine-tuned) and MinkUNet+ViT (frozen probing, FOV/Full mIoU).}
\label{tab:semkitti}
\setlength{\tabcolsep}{6pt}
\begin{tabular}{l c cc}
\toprule
& PMF & \multicolumn{2}{c}{MinkUNet+ViT} \\
\cmidrule(lr){2-2}\cmidrule(lr){3-4}
Pretrain & mIoU & FOV mIoU & Full mIoU \\
\midrule
Random & 63.9 & 31.1 & 31.8 \\
DeCUR & \underline{65.0} & \underline{44.8} & \textbf{44.4} \\
DenseTwins & \textbf{65.2} & \textbf{47.1} & \underline{44.1} \\
\bottomrule
\end{tabular}
\end{table}

We evaluate the RGB--LiDAR setting on Semantic-KITTI~\cite{behley2019iccv}, both by fine-tuning PMF~\cite{zhuang2021perception} and by using frozen backbones. For the latter, we use MinkUNet~\cite{choy2019minkowski} as 3D LiDAR backbone, which we fuse with ViT RGB features as in PointFusion~\cite{xu2018pointfusion}. We report in \Cref{tab:semkitti}, for frozen-backbone probing, the mIoU both in the region of the 3D scan inside the camera field of view (FOV mIoU) and in the full 360\textdegree{} scan (mIoU). DenseTwins leads in FOV mIoU, where RGB features are available for every point, while DeCUR is marginally better in mIoU, where most points fall outside the camera field of view and therefore lack RGB features, suggesting that DenseTwins' frozen representations are particularly effective where both modalities are actually available. DenseTwins improves on DeCUR also when fine-tuning.

\subsection{Ablation Studies}

\begin{table}[t]
\centering

\begin{minipage}{0.51\linewidth}
\centering
\caption{Effect of the global and dense projectors on SUN-RGBD.}
\label{tab:ablation-projector}
\small
\setlength{\tabcolsep}{4pt}
\resizebox{\linewidth}{!}{%
\begin{tabular}{cccc}
\toprule
Global Proj. & Dense Proj. & mIoU & LP mIoU \\
\midrule
\xmark & \xmark & 50.7 & 28.3 \\
\xmark & \cmark & 50.9 & 29.1 \\
\cmark & \xmark & 51.1 & 32.0 \\
\cmark & \cmark & \textbf{51.5} & \textbf{33.7} \\
\bottomrule
\end{tabular}
}
\end{minipage}
\hfill
\begin{minipage}{0.45\linewidth}
\centering
\caption{Effect of the loss components on SUN-RGBD.}
\label{tab:ablation-losses}
\small
\setlength{\tabcolsep}{3pt}
\resizebox{\linewidth}{!}{%
\begin{tabular}{ccccrr}
\toprule
$\mathcal{L}_{g}^{m}$ &
$\mathcal{L}_{d}^{m}$ &
$\mathcal{L}_{\text{cross}}^{g}$ &
$\mathcal{L}_{\text{cross}}^{d}$ &
mIoU & LP mIoU \\
\midrule
\cmark & \xmark & \cmark & \xmark & 50.6 & 23.7 \\
\xmark & \cmark & \xmark & \cmark & 51.3 & 25.3 \\
\xmark & \cmark & \cmark & \cmark & 51.4 & 31.0 \\
\cmark & \cmark & \cmark & \cmark & \textbf{51.5} & \textbf{33.7} \\
\bottomrule
\end{tabular}
}
\end{minipage}

\end{table}

\begin{figure}[t]
\centering

\begin{minipage}{0.49\linewidth}
    \centering
    \includegraphics[width=\linewidth]{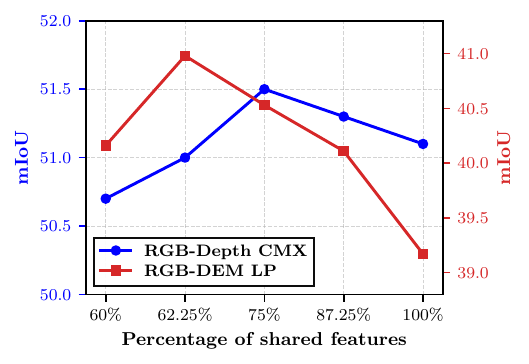}
    \caption{Effect of the percentage of shared features on SUN-RGBD and GeoNRW.}
    \label{fig:ablation-dim}
\end{minipage}
\hfill
\begin{minipage}{0.49\linewidth}
    \centering
    \includegraphics[width=\linewidth]{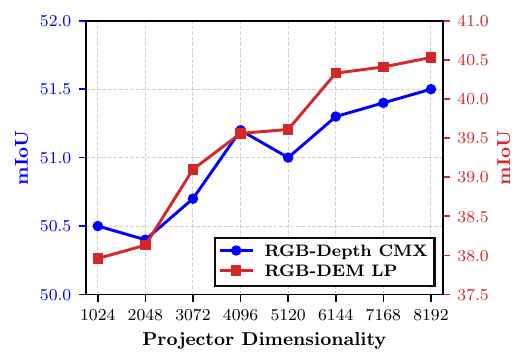}
    \caption{Effect of the dimensionality of $K_p$, on SUN-RGBD and GeoNRW.}
    \label{fig:ablation-dim-proj}
\end{minipage}

\end{figure}

We perform a series of ablation studies to analyze the impact of the main design choices of our framework, including the dimensionality of the projection heads, the percentage of modality-shared features, and the role of the global and dense objectives, using SUN-RGBD~\cite{song2015sun} and, for the first two ablations, also GeoNRW~\cite{geonrw}.\\
\textbf{Projector dimensionality.}
We first analyze the effect of the projector dimensionality $K_p$. As shown in \Cref{fig:ablation-dim-proj}, increasing $K_p$ generally improves performance, suggesting that larger projection spaces allow the model to better capture multimodal correspondences.\\
\textbf{Decoupling percentage.}
We then investigate the ratio $K_s/K_p$ of modality-shared components. As shown in \Cref{fig:ablation-dim}, the optimum varies across modality pairs: $K_s \approx 75\%,K_p$ for RGB-Depth and $K_s \approx 62\%,K_p$ for GeoNRW. Too few shared components limit cross-modal alignment, whereas sharing all features degrades performance by preventing the model from preserving modality-specific information. Automatically adapting this split to each modality pair is left for future work.\\
\textbf{Effect of the projector heads.}
We further study the contribution of the global and dense projection heads. As shown in \Cref{tab:ablation-projector}, both components positively contribute to performance, under both the fine-tuned CMX mIoU and the frozen linear-probe (LP) mIoU on SUN-RGBD. The global projector contributes more individually (+0.4 mIoU) than the dense one (+0.2 mIoU), yet their combination yields a larger joint gain (+0.8 mIoU), confirming complementary roles; the same trend holds under linear probing, with an even larger gap (+5.4 LP mIoU). This benefit comes at a moderate computational cost, quantified in \Cref{sec:supp-computational-overhead}.\\
\textbf{Effect of the loss components.}
Finally, we analyze the impact of the different training loss terms. As reported in \Cref{tab:ablation-losses}, both the global and dense objectives contribute to the final performance, with the dense losses ($\mathcal{L}_{d}^{(m)}$ and $\mathcal{L}_{\text{cross}}^{d}$) providing the largest gain. In particular, training with dense objectives only already outperforms the global-only configuration by $+0.7$ mIoU (51.3 vs.\ 50.6), confirming that dense supervision is the core contribution of DenseTwins; adding the global objectives further improves performance under both fine-tuning and linear probing, showing that the two scales of supervision are complementary. We further verify in \Cref{sec:supp-dense-matching-robustness} that the dense correspondences underlying $\mathcal{L}_{d}^{(m)}$ and $\mathcal{L}_{\text{cross}}^{d}$ remain robust under moderate spatial misalignment between modalities.

Beyond the quantitative ablations presented in this section, we provide additional analyses of the learned shared and modality-specific subspaces in the supplementary material to verify that DenseTwins successfully disentangles shared and modality-specific information (\Cref{sec:supp-analysis}).

\subsection{Discussion}

A consistent trend across the experiments is that contrastive methods such as SimCLR and CLIP remain among the strongest baselines, often outperforming redundancy-reduction approaches such as Barlow Twins, VICReg, and even DeCUR. This suggests that contrastive learning may be worth exploring for aligning the shared subspace in place of redundancy reduction, and indicates that dense correspondences play an important role that these methods lack.

DenseTwins nonetheless achieves the best overall results across modality pairs, application domains, and evaluation protocols, from fine-tuning to frozen-backbone probing and object detection, indicating that dense modality decoupling captures information complementary to existing self-supervised objectives, rather than replacing them.

\section{Conclusion}

In this paper, we presented \textbf{DenseTwins}, a multimodal self-supervised framework for learning dense representations while explicitly separating modality-shared and modality-specific information. 
DenseTwins extends redundancy-reduction and modality-decoupling principles from global embeddings to dense token-level representations by establishing dense correspondences and combining intra-modal and cross-modal objectives.
Through experiments across multiple sensing setups, we showed that DenseTwins consistently improves over existing self-supervised baselines on downstream tasks. 
These results demonstrate the effectiveness of dense modality decoupling for learning transferable multimodal representations.

\section*{Acknowledgements}
This paper is supported by  FAIR  (NextGenerationEU program, PNRR-PE-AI scheme, M4C2, Investment 1.3, Line on Artificial Intelligence) and by PEPR IA on grant SHARP ANR-23-PEIA-0008. We acknowledge the EuroHPC Joint Undertaking for awarding this project access to the EuroHPC supercomputer Karolina hosted by IT4Innovations, Czech Republic through an EuroHPC Development Access call. Part of this work was granted access to the JeanZay HPC/AI resources of IDRIS under the allocation AD011017661.

{
    \small
    \bibliographystyle{ieeenat_fullname}
    \bibliography{main}

@String(CVPR= {IEEE Conf. Comput. Vis. Pattern Recog.})

@String(ICCV= {Int. Conf. Comput. Vis.})

@String(ECCV= {Eur. Conf. Comput. Vis.})

@String(CVPR  = {CVPR})

@String(ICCV  = {ICCV})

@String(ECCV  = {ECCV})

@inproceedings{chen2020simclr,
  title        = {A Simple Framework for Contrastive Learning of Visual Representations},
  author       = {Ting Chen and Simon Kornblith and Mohammad Norouzi and Geoffrey Hinton},
  booktitle    = {International Conference on Machine Learning (ICML)},
  year         = {2020},
  url          = {https://arxiv.org/abs/2002.05709}
}

@article{he2020moco,
  title        = {Momentum Contrast for Unsupervised Visual Representation Learning},
  author       = {Kaiming He and Haoqi Fan and Yuxin Wu and Saining Xie and Ross Girshick},
  journal      = {arXiv preprint arXiv:1911.05722},
  year         = {2020},
  url          = {https://arxiv.org/abs/1911.05722}
}

@article{he2020mocov2,
  title        = {MoCo v2: Improved Baselines with Momentum Contrastive Learning},
  author       = {Kaiming He and Haoqi Fan and Yuxin Wu and Saining Xie and Ross Girshick},
  journal      = {arXiv preprint arXiv:2003.04297},
  year         = {2020},
  url          = {https://arxiv.org/abs/2003.04297}
}

@inproceedings{zbontar2021barlow,
  title        = {Barlow Twins: Self-Supervised Learning via Redundancy Reduction},
  author       = {Jure Zbontar and Li Jing and Ishan Misra and Yann LeCun and Stephane Deny},
  booktitle    = {Proceedings of the 38th International Conference on Machine Learning (ICML)},
  volume       = {139},
  pages        = {12310--12320},
  year         = {2021},
  url          = {https://proceedings.mlr.press/v139/zbontar21a.html}
}

@inproceedings{wang2021densecl,
  title        = {Dense Contrastive Learning for Self-Supervised Visual Pre-Training},
  author       = {Xinlong Wang and Rufeng Zhang and Chunhua Shen and Tao Kong and Lei Li},
  booktitle    = {Proceedings of the IEEE/CVF Conference on Computer Vision and Pattern Recognition (CVPR)},
  year         = {2021},
  url          = {https://arxiv.org/abs/2011.09157}
}

@inproceedings{wang2024decur,
  title        = {Decoupling Common and Unique Representations for Multimodal Self-Supervised Learning (DeCUR)},
  author       = {Yi Wang and Conrad M. Albrecht and Nassim Ait Ali Braham and Chenying Liu and Zhitong Xiong and Xiao Xiang Zhu},
  booktitle    = {Computer Vision – ECCV 2024, Lecture Notes in Computer Science, vol. 15087},
  year         = {2024},
  publisher    = {Springer},
  url          = {https://arxiv.org/abs/2309.05300}
}

@inproceedings{hwang2015kaist,
  title        = {Multispectral Pedestrian Detection: Benchmark Dataset and Baselines},
  author       = {Soonmin Hwang and Jaesik Park and Namil Kim and Yukyung Choi and In So Kweon},
  booktitle    = {Proceedings of the IEEE/CVF Conference on Computer Vision and Pattern Recognition (CVPR)},
  pages        = {1037--1045},
  year         = {2015},
  organization = {IEEE},
  url          = {https://ieeexplore.ieee.org/document/7298684},
  doi          = {10.1109/CVPR.2015.7298684}
}

@inproceedings{he2016resnet,
  title        = {Deep Residual Learning for Image Recognition},
  author       = {Kaiming He and Xiangyu Zhang and Shaoqing Ren and Jian Sun},
  booktitle    = {Proceedings of the IEEE Conference on Computer Vision and Pattern Recognition (CVPR)},
  pages        = {770--778},
  year         = {2016},
  organization = {IEEE},
  url          = {https://arxiv.org/abs/1512.03385},
  doi          = {10.1109/CVPR.2016.90}
}

@article{iskender2022improving,
  title={Improving dense contrastive learning with dense negative pairs},
  author={Iskender, Berk and Xu, Zhenlin and Kornblith, Simon and Chu, En-Hung and Khademi, Maryam},
  journal={arXiv preprint arXiv:2210.05063},
  year={2022}
}

@article{bardes2021vicreg,
  title={Vicreg: Variance-invariance-covariance regularization for self-supervised learning},
  author={Bardes, Adrien and Ponce, Jean and LeCun, Yann},
  journal={arXiv preprint arXiv:2105.04906},
  year={2021}
}

@inproceedings{radford2021learning,
  title={Learning transferable visual models from natural language supervision},
  author={Radford, Alec and Kim, Jong Wook and Hallacy, Chris and Ramesh, Aditya and Goh, Gabriel and Agarwal, Sandhini and Sastry, Girish and Askell, Amanda and Mishkin, Pamela and Clark, Jack and others},
  booktitle={International conference on machine learning},
  pages={8748--8763},
  year={2021},
  organization={PmLR}
}

@inproceedings{xie2021propagate,
  title={Propagate yourself: Exploring pixel-level consistency for unsupervised visual representation learning},
  author={Xie, Zhenda and Lin, Yutong and Zhang, Zheng and Cao, Yue and Lin, Stephen and Hu, Han},
  booktitle={Proceedings of the IEEE/CVF conference on computer vision and pattern recognition},
  pages={16684--16693},
  year={2021}
}

@article{zhang2023cmx,
  title={CMX: Cross-modal fusion for RGB-X semantic segmentation with transformers},
  author={Zhang, Jiaming and Liu, Huayao and Yang, Kailun and Hu, Xinxin and Liu, Ruiping and Stiefelhagen, Rainer},
  journal={IEEE Transactions on intelligent transportation systems},
  volume={24},
  number={12},
  pages={14679--14694},
  year={2023},
  publisher={IEEE}
}

@inproceedings{zhuang2021perception,
  title={Perception-aware multi-sensor fusion for 3d lidar semantic segmentation},
  author={Zhuang, Zhuangwei and Li, Rong and Jia, Kui and Wang, Qicheng and Li, Yuanqing and Tan, Mingkui},
  booktitle={Proceedings of the IEEE/CVF international conference on computer vision},
  pages={16280--16290},
  year={2021}
}

@inproceedings{song2015sun,
  title={Sun rgb-d: A rgb-d scene understanding benchmark suite},
  author={Song, Shuran and Lichtenberg, Samuel P and Xiao, Jianxiong},
  booktitle={Proceedings of the IEEE conference on computer vision and pattern recognition},
  pages={567--576},
  year={2015}
}

@Article{RAL21Gehrig,
  author        = {Daniel Gehrig and Michelle Rüegg and Mathias Gehrig and Javier Hidalgo-Carrio and Davide Scaramuzza},
  title         = {Combining Events and Frames using Recurrent Asynchronous Multimodal Networks for Monocular Depth Prediction},
  journal       = {{IEEE} Robotic and Automation Letters. (RA-L)},
  url           = {http://rpg.ifi.uzh.ch/docs/RAL21_Gehrig.pdf},
  year          = {2021}
}

@data{geonrw,
doi = {10.21227/s5xq-b822},
url = {https://dx.doi.org/10.21227/s5xq-b822},
author = {Gerald Baier and Antonin Deschemps and Michael Schmitt and Naoto Yokoya},
publisher = {IEEE Dataport},
title = {GeoNRW},
year = {2020} }

@inproceedings{behley2019iccv,
  author = {J. Behley and M. Garbade and A. Milioto and J. Quenzel and S. Behnke and C. Stachniss and J. Gall},
  title = {{SemanticKITTI: A Dataset for Semantic Scene Understanding of LiDAR Sequences}},
  booktitle = {Proc. of the IEEE/CVF International Conf.~on Computer Vision (ICCV)},
  year = {2019}
}

@article{xie2021segformer,
  title={SegFormer: Simple and efficient design for semantic segmentation with transformers},
  author={Xie, Enze and Wang, Wenhai and Yu, Zhiding and Anandkumar, Anima and Alvarez, Jose M and Luo, Ping},
  journal={Advances in neural information processing systems},
  volume={34},
  pages={12077--12090},
  year={2021}
}

@inproceedings{long2015fully,
  title={Fully convolutional networks for semantic segmentation},
  author={Long, Jonathan and Shelhamer, Evan and Darrell, Trevor},
  booktitle={Proceedings of the IEEE conference on computer vision and pattern recognition},
  pages={3431--3440},
  year={2015}
}

@article{zong2024self,
  title={Self-supervised multimodal learning: A survey},
  author={Zong, Yongshuo and Mac Aodha, Oisin and Hospedales, Timothy M},
  journal={IEEE Transactions on Pattern Analysis and Machine Intelligence},
  volume={47},
  number={7},
  pages={5299--5318},
  year={2024},
  publisher={IEEE}
}

@article{sgaravatti2026lcf3d,
  title={LCF3D: A Robust and Real-Time Late-Cascade Fusion Framework for 3D Object Detection in Autonomous Driving},
  author={Sgaravatti, Carlo and Pieroni, Riccardo and Corno, Matteo and Savaresi, Sergio M and Magri, Luca and Boracchi, Giacomo},
  journal={Pattern Recognition},
  pages={113046},
  year={2026},
  publisher={Elsevier}
}

@ARTICLE{8269806,
  author={Baltrušaitis, Tadas and Ahuja, Chaitanya and Morency, Louis-Philippe},
  journal={IEEE Transactions on Pattern Analysis and Machine Intelligence}, 
  title={Multimodal Machine Learning: A Survey and Taxonomy}, 
  year={2019},
  volume={41},
  number={2},
  pages={423-443},
  doi={10.1109/TPAMI.2018.2798607}}

@article{bayoudh2024survey,
  title={A survey of multimodal hybrid deep learning for computer vision: Architectures, applications, trends, and challenges},
  author={Bayoudh, Khaled},
  journal={Information Fusion},
  volume={105},
  pages={102217},
  year={2024},
  publisher={Elsevier}
}

@inproceedings{gupta2014learning,
  title={Learning rich features from RGB-D images for object detection and segmentation},
  author={Gupta, Saurabh and Girshick, Ross and Arbel{\'a}ez, Pablo and Malik, Jitendra},
  booktitle={European conference on computer vision},
  pages={345--360},
  year={2014},
  organization={Springer}
}

@article{sumbul2019bigearthnet,
  title={Bigearthnet: A large-scale benchmark archive for remote sensing image understanding},
  author={Sumbul, Gencer and Charfuelan, Marcela and Demir, Beg{\"u}m and Markl, Volker},
  journal={arXiv preprint arXiv:1902.06148},
  year={2019}
}

@inproceedings{caron2021emerging,
  title={Emerging properties in self-supervised vision transformers},
  author={Caron, Mathilde and Touvron, Hugo and Misra, Ishan and J{\'e}gou, Herv{\'e} and Mairal, Julien and Bojanowski, Piotr and Joulin, Armand},
  booktitle={Proceedings of the IEEE/CVF international conference on computer vision},
  pages={9650--9660},
  year={2021}
}

@article{oquab2023dinov2,
  title={Dinov2: Learning robust visual features without supervision},
  author={Oquab, Maxime and Darcet, Timoth{\'e}e and Moutakanni, Th{\'e}o and Vo, Huy and Szafraniec, Marc and Khalidov, Vasil and Fernandez, Pierre and Haziza, Daniel and Massa, Francisco and El-Nouby, Alaaeldin and others},
  journal={arXiv preprint arXiv:2304.07193},
  year={2023}
}

@inproceedings{assran2023self,
  title={Self-supervised learning from images with a joint-embedding predictive architecture},
  author={Assran, Mahmoud and Duval, Quentin and Misra, Ishan and Bojanowski, Piotr and Vincent, Pascal and Rabbat, Michael and LeCun, Yann and Ballas, Nicolas},
  booktitle={Proceedings of the IEEE/CVF conference on computer vision and pattern recognition},
  pages={15619--15629},
  year={2023}
}

@article{zhou2021ibot,
  title={ibot: Image bert pre-training with online tokenizer},
  author={Zhou, Jinghao and Wei, Chen and Wang, Huiyu and Shen, Wei and Xie, Cihang and Yuille, Alan and Kong, Tao},
  journal={arXiv preprint arXiv:2111.07832},
  year={2021}
}

@article{balestriero2025lejepa,
  title={Lejepa: Provable and scalable self-supervised learning without the heuristics},
  author={Balestriero, Randall and LeCun, Yann},
  journal={arXiv preprint arXiv:2511.08544},
  year={2025}
}

@article{Hazarika2020MISAMA,
  title={MISA: Modality-Invariant and -Specific Representations for Multimodal Sentiment Analysis},
  author={Devamanyu Hazarika and Roger Zimmermann and Soujanya Poria},
  journal={Proceedings of the 28th ACM International Conference on Multimedia},
  year={2020},
  url={https://api.semanticscholar.org/CorpusID:218538102}
}

@article{Lai2023SharedAP,
  title={Shared and Private Information Learning in Multimodal Sentiment Analysis with Deep Modal Alignment and Self-supervised Multi-Task Learning},
  author={Songning Lai and Jiakang Li and Guinan Guo and Xifeng Hu and Yulong Li and Yuan Tan and Zichen Song and Yutong Liu and Zhaoxia Ren and Zhi Liu and Chun Wang and Danmin Miao},
  journal={2024 International Joint Conference on Neural Networks (IJCNN)},
  year={2023},
  pages={1-8},
  url={https://api.semanticscholar.org/CorpusID:258686191}
}

@article{alayrac2020self,
  title={Self-supervised multimodal versatile networks},
  author={Alayrac, Jean-Baptiste and Recasens, Adria and Schneider, Rosalia and Arandjelovi{\'c}, Relja and Ramapuram, Jason and De Fauw, Jeffrey and Smaira, Lucas and Dieleman, Sander and Zisserman, Andrew},
  journal={Advances in neural information processing systems},
  volume={33},
  pages={25--37},
  year={2020}
}

@inproceedings{hao2021cross,
  title={Cross-modality person re-identification via modality confusion and center aggregation},
  author={Hao, Xin and Zhao, Sanyuan and Ye, Mang and Shen, Jianbing},
  booktitle={2021 IEEE/CVF International Conference on Computer Vision (ICCV)},
  pages={16383--16392},
  year={2021},
  organization={IEEE}
}

@article{wang2024information,
  title={An information criterion for controlled disentanglement of multimodal data},
  author={Wang, Chenyu and Gupta, Sharut and Zhang, Xinyi and Tonekaboni, Sana and Jegelka, Stefanie and Jaakkola, Tommi and Uhler, Caroline},
  journal={arXiv preprint arXiv:2410.23996},
  year={2024}
}

@article{gehrig2021dsec,
  title={Dsec: A stereo event camera dataset for driving scenarios},
  author={Gehrig, Mathias and Aarents, Willem and Gehrig, Daniel and Scaramuzza, Davide},
  journal={IEEE Robotics and Automation Letters},
  volume={6},
  number={3},
  pages={4947--4954},
  year={2021},
  publisher={IEEE}
}

@inproceedings{tian2019fcos,
  title={Fcos: Fully convolutional one-stage object detection},
  author={Tian, Zhi and Shen, Chunhua and Chen, Hao and He, Tong},
  booktitle={2019 IEEE/CVF international conference on computer vision (ICCV)},
  pages={9626--9635},
  year={2019},
  organization={Ieee}
}

@inproceedings{choy2019minkowski,
  title={4d spatio-temporal convnets: Minkowski convolutional neural networks},
  author={Choy, Christopher and Gwak, JunYoung and Savarese, Silvio},
  booktitle={2019 IEEE/CVF conference on computer vision and pattern recognition (CVPR)},
  pages={3070--3079},
  year={2019},
  organization={IEEE}
}

@inproceedings{xu2018pointfusion,
  title={Pointfusion: Deep sensor fusion for 3d bounding box estimation},
  author={Xu, Danfei and Anguelov, Dragomir and Jain, Ashesh},
  booktitle={2018 IEEE/CVF conference on computer vision and pattern recognition},
  pages={244--253},
  year={2018},
  organization={IEEE}
}

@article{couprie2013indoor,
  title={Indoor semantic segmentation using depth information},
  author={Couprie, Camille and Farabet, Cl{\'e}ment and Najman, Laurent and LeCun, Yann},
  journal={arXiv preprint arXiv:1301.3572},
  year={2013}
}
}

\clearpage
\setcounter{page}{1}
\maketitlesupplementary

In this document, we provide additional details about the implementation of DenseTwins for the multimodal datasets (SUN-RGBD, KAIST, DSEC, EventScape, GeoNRW, SemanticKITTI and BigEarthNet) used for testing. Additionally, we show the pseudocode of the training procedure of DenseTwins, an analysis of the learned shared/specific decomposition, an analysis of the computational overhead introduced by the projection heads and the dense matching mechanism, and some visualizations of the downstream results of CMX \cite{zhang2023cmx} on SUN-RGBD and EventScape.

\appendix

\section{Implementation Details}
\label{sec:implementation-details-supp}

\subsection{Pretraining}
\label{sec:supp-pretraining}

We pretrain multimodal backbones using DenseTwins on seven multimodal datasets covering different sensing modalities: RGB--Depth (SUN-RGBD \cite{song2015sun}), RGB--Event (EventScape \cite{RAL21Gehrig}, DSEC \cite{gehrig2021dsec}), RGB--Thermal (KAIST \cite{hwang2015kaist}), RGB--DEM (GeoNRW \cite{geonrw}), RGB--LiDAR (Semantic-KITTI \cite{behley2019iccv}), and SAR--Optical (BigEarthNet \cite{sumbul2019bigearthnet}).
All pretraining experiments are run on $8\times$A100 GPUs; unless otherwise noted, batch sizes are reported per GPU.
For each dataset, two independently augmented views are generated per modality following standard self-supervised learning practices. Augmentations are applied independently to each modality to generate two views per input pair.

\paragraph{SUN-RGBD (RGB--Depth).}
SUN-RGBD consists of indoor RGB images paired with depth maps captured using different RGB-D sensors.
Following common practice in RGB-D scene understanding, the depth maps are preprocessed into the HHA representation \cite{gupta2014learning}, which encodes horizontal disparity, height above ground, and angle with gravity.
Following DeCUR~\cite{wang2024decur}, for the fine-tuning protocol both RGB and HHA images are resized using random resized cropping to a spatial resolution of $224 \times 224$ pixels, using the same augmentations as DeCUR: random resized cropping, horizontal flipping, color jittering and Gaussian blur for RGB images, and geometric augmentations only (random cropping and horizontal flipping) for HHA images.
For this protocol we adopt MiT-B2 encoders following the SegFormer architecture \cite{xie2021segformer}, with a projector composed of two convolutional layers with kernel size 1.
Pretraining is performed using the AdamW optimizer with learning rate $1 \times 10^{-5}$, batch size 96, for 200 epochs.
For the frozen-backbone probing protocol, we instead pretrain a ViT-S/14 backbone (14-pixel patch size) at the original image resolution ($640 \times 480$), using the same augmentations, with batch size 64 for 200 epochs.

\paragraph{EventScape (RGB--Event).}
EventScape provides synchronized RGB frames and event streams captured in driving scenarios.
Event streams are converted into voxel grid representations by discretizing the temporal dimension into three bins.
The resulting voxel grids are stored as three-channel images, allowing them to be processed similarly to RGB inputs.
Both RGB images and event voxel grids are resized using random resized cropping to $224 \times 224$ pixels.
RGB images are augmented using random resized cropping, horizontal flipping, color jittering and Gaussian blur, while event voxel grids undergo geometric augmentations such as random cropping and horizontal flipping.
Pretraining is performed using the AdamW optimizer with learning rate $1 \times 10^{-5}$, batch size 96, for 50 epochs.

\paragraph{KAIST (RGB--Thermal).}
KAIST is a multimodal pedestrian-detection benchmark for autonomous driving, providing RGB images paired with thermal images and object-detection annotations for pedestrians.
Thermal images are single-channel (grayscale); we repeat the channel three times to obtain a 3-channel input.
We pretrain ViT-S/16, Swin-T, and ConvNeXtV2-T backbones at the original image resolution ($640 \times 480$), using the AdamW optimizer with learning rate $1 \times 10^{-4}$, batch size 64, for 100 epochs.
For these backbones, the projector is a 3-layer vision transformer, following the predictor design of I-JEPA~\cite{assran2023self}.

\paragraph{DSEC (RGB--Event).}
DSEC is a multimodal object-detection benchmark for autonomous driving, providing RGB images paired with event streams.
As in EventScape, events are discretized into a voxel grid, here using 5 temporal bins, yielding a 5-channel input at the original image resolution ($640 \times 480$); this changes, e.g., the patch embedding of the ViT backbone to accept 5-channel inputs.
Pretraining otherwise follows the same protocol as KAIST: ViT-S/16, Swin-T, and ConvNeXtV2-T backbones, learning rate $1 \times 10^{-4}$, batch size 64, for 100 epochs.

\paragraph{GeoNRW (RGB--DEM).}
GeoNRW is a multimodal remote sensing dataset composed of aerial RGB orthophotos and digital elevation models (DEM).
Since the original scenes are very large, we generate training samples by extracting tiles of size $250 \times 250$ pixels from the raw images.
DEM maps are normalized to the $[0,1]$ range and resized to match the RGB spatial resolution.
Both modalities are resized using random resized cropping to $224 \times 224$ pixels during training.
Augmentations include random cropping and horizontal flipping for both modalities, while color jittering is applied only to RGB images.
For this fine-tuning protocol we adopt ResNet50 encoders \cite{he2016resnet}.
Pretraining is performed using the AdamW optimizer with learning rate $1 \times 10^{-5}$, batch size 96, for 50 epochs.
For the frozen-backbone probing protocol, we instead pretrain a ViT-S/16 backbone for 50 epochs, using the same preprocessing and image resolution.

\paragraph{SemanticKITTI (RGB--LiDAR).}
SemanticKITTI provides outdoor driving sequences containing LiDAR point clouds and synchronized camera images.
Since camera images are available only for the front field of view, the LiDAR point clouds are spatially cropped to match the camera field of view before being processed by the model.
The cropped point clouds are then converted into dense representations used by the LiDAR encoder.
RGB images are resized using random resized cropping to $224 \times 224$ pixels and augmented with horizontal flipping and color jittering.
LiDAR representations undergo geometric augmentations consistent with the projection operation.
For this fine-tuning protocol we adopt a ResNet34 encoder for the camera branch and a VoxelNet-style encoder for the LiDAR branch.
Pretraining is performed using the AdamW optimizer with learning rate $1 \times 10^{-6}$, batch size 32, for 200 epochs.
For the frozen-backbone probing protocol, we instead pretrain a MinkUNet~\cite{choy2019minkowski} encoder for the LiDAR branch and a ViT backbone for the camera branch, using the AdamW optimizer with learning rate $1 \times 10^{-5}$, batch size 32, for 200 epochs.

\paragraph{BigEarthNet (SAR--Optical).}
For the BigEarthNet benchmark, we follow the multimodal setting introduced in DeCUR and use Sentinel-1 SAR and Sentinel-2 optical imagery as the two modalities. Sentinel-1 images contain two channels (VV and VH polarization), while Sentinel-2 images contain 13 multispectral bands.
Following DeCUR, all images are resampled to a common spatial resolution of 10\,m per pixel. Optical channels are normalized using band-wise statistics, while SAR images are clipped to remove extreme outliers and subsequently normalized.
We use a ResNet50 backbone for both modalities. Optimization follows the same protocol adopted for GeoNRW, using AdamW with learning rate $10^{-5}$, weight decay $10^{-4}$, cosine learning-rate decay, and batch size 96 for 100 epochs.

\paragraph{Hyperparameters}
Across all datasets, the encoders are followed by projection heads that map the features into a $K_p$-dimensional embedding space used during self-supervised training.
Following common practice in redundancy-reduction-based self-supervised learning methods, we use a high-dimensional embedding space with $K_p=8192$. The embedding space is divided into shared and modality-specific subspaces, with $K_s$ shared dimensions and $K_u$ modality-specific dimensions ($K_p = K_s + K_u$).
The projection heads are used only during pretraining and are discarded when transferring the encoders to downstream tasks.

\subsection{Downstream Evaluation Protocols}
\label{sec:supp-finetuning}

To evaluate the quality of the learned representations, the pretrained encoders are transferred to downstream tasks under the two protocols introduced in \Cref{sec:eval-protocols}: fine-tuning and frozen-backbone probing.

\paragraph{Frozen-Backbone Probing.}
For semantic segmentation (SUN-RGBD, GeoNRW), we evaluate the frozen backbones with three protocols: k-nearest-neighbor classification with $k=10$ neighbors retrieved among the patch tokens of the training set; linear probing, training a linear classifier on the patch tokens for 20 epochs; and a lightweight FCN head, trained for 30 epochs on the feature maps of the last three ViT stages.
For object detection (KAIST, DSEC), we attach a FCOS~\cite{tian2019fcos} head to the last three feature maps of the frozen backbone, combined through a feature pyramid network, and train it for 20 epochs.
For SemanticKITTI, we train the downstream model on the full $360$\textdegree{} scan, fusing the frozen MinkUNet and ViT backbones by projecting each 3D point onto the image plane and concatenating the corresponding RGB feature to the point feature, as in PointFusion~\cite{xu2018pointfusion}. Points falling outside the camera field of view receive a zero feature vector. The fused features are then processed by the MinkUNet decoder, which serves as the segmentation head and is trained for 20 epochs while both backbones remain frozen. We report mIoU both within the camera field of view and over the entire scan.

\paragraph{Low-Label Setting.}
For the low-label fine-tuning experiments (SUN-RGBD, $10\%$ labels; GeoNRW, $1\%$ labels), the labelled subset of the training set is drawn once via simple random sampling, using a fixed random seed shared across all compared methods, so that every method is fine-tuned on the exact same subset; validation and test sets are left unchanged. The frozen-backbone probing protocol, in contrast, always uses the full training set.

\paragraph{SUN-RGBD (Fine-Tuning).}
For SUN-RGBD we adopt the CMX \cite{zhang2023cmx} architecture for RGB-D semantic segmentation.
The pretrained RGB and depth encoders are used to initialize the corresponding branches of the model.
Training follows the same configuration as the original CMX paper and runs for 200 epochs.

\paragraph{EventScape (Fine-Tuning).}
For EventScape we also use the CMX architecture to fuse RGB frames and event representations.
The pretrained encoders are used to initialize the modality-specific branches and the model is trained for 50 epochs using the same training configuration as the original CMX implementation.

\paragraph{GeoNRW (Fine-Tuning).}
For GeoNRW we evaluate the learned representations using a lightweight fully convolutional segmentation model \cite{long2015fully}.
The feature maps extracted from the two modality-specific encoders are concatenated and passed to a segmentation head composed of upsampling layers followed by a linear classification layer.
Specifically, we use feature maps from the last three stages of the ResNet backbone and apply three different upsampling factors before combining them to produce the final prediction map.
The segmentation model is trained for 30 epochs.

\paragraph{SemanticKITTI (Fine-Tuning).}
For SemanticKITTI, we adopt the PMF \cite{zhuang2021perception} architecture for LiDAR-camera semantic segmentation.
We follow exactly the same training protocol as the original PMF implementation and train the model for 50 epochs.
To ensure compatibility with the architecture, we load the pretrained weights only for the camera branch of the model while keeping the LiDAR backbone unchanged, as it requires RGB features to be fused in an early stage of the LiDAR backbone, which is not compatible with our pretraining scheme.

\paragraph{BigEarthNet.}
For BigEarthNet, we evaluate the learned representations on multilabel land-use classification. Following a standard linear probing protocol, the pretrained encoder is kept frozen, and a linear classification head is trained on top of the extracted features. The classification head consists of a single fully-connected layer mapping the backbone features to the target classes. Performance is evaluated using mean Average Precision (mAP).

\subsection{Baseline Methods}
\label{sec:supp-competitors}

We compare the proposed method with contrastive, redundancy-reduction-based, and dense single-modal self-supervised learning baselines, including SimCLR \cite{chen2020simclr}, VICReg \cite{bardes2021vicreg}, Barlow Twins \cite{zbontar2021barlow}, CLIP \cite{radford2021learning}, DeCUR \cite{wang2024decur}, DINO \cite{caron2021emerging}, iBOT \cite{zhou2021ibot}, I-JEPA \cite{assran2023self}, and LeJEPA \cite{balestriero2025lejepa}. All baselines are trained using the same backbones, data augmentations, and optimization settings whenever applicable.

SimCLR, VICReg, and Barlow Twins are originally designed for single-modal representation learning.
In our multimodal setting, we adapt these methods by treating the two modalities as the two views used during training. In practice, to train all three methods, we use one augmented view per modality and apply the loss of SimCLR, VICReg, or Barlow Twins to the embeddings extracted from the two modalities instead of two different augmented views of the same modality, allowing the methods to learn cross-modal representations.

For CLIP, we use the implementation provided in the DeCUR codebase to ensure consistent training settings across baselines.

For DeCUR, we use the official implementation released by the authors, adapting only the dataset loading and preprocessing pipelines to match the datasets used in our experiments.

For DINO, iBOT, I-JEPA, and LeJEPA, we use the official code released by the respective authors, adopting the same ViT backbone as DenseTwins but each method's own official hyperparameters for augmentations and image sizes, except for the number of training epochs and batch sizes, which we set to match our own pretraining schedule. Each method is trained independently on each modality, producing two single-modal backbones that are combined via late fusion for downstream evaluation, as described in \Cref{sec:eval-protocols}.

\section{Algorithm Pseudocode}

In this section we summarize the training procedure of DenseTwins. \Cref{alg:densetwins} describes the overall training loop, while \Cref{alg:btloss,alg:uniqueloss} define the two loss functions used to optimize the model.

Given a pair of observations from two modalities, two independently augmented views are generated for each modality. The augmented samples are processed by two modality-specific encoders followed by projection heads that produce both global embeddings and dense token-level representations. The training objective combines four components: intra-modal global losses, intra-modal dense losses, cross-modal global losses and cross-modal dense losses.

\begin{algorithm}[h]
\caption{Barlow Twins Loss}
\label{alg:btloss}
\begin{algorithmic}[1]
\Require Cross-correlation matrix $C$, coefficient $\lambda$
\Function{BarlowTwinsLoss}{$C$, $\lambda$}
\State $l_{\text{on-diag}} = \sum_{p} (1 - C_{pp})^2$ 
\State $l_{\text{off-diag}} = \sum_{p \neq q} (C_{pq})^2$
\State \Return $l_{\text{on-diag}} + \lambda \cdot l_{\text{off-diag}}$
\EndFunction
\end{algorithmic}
\end{algorithm}

\begin{algorithm}[h]
\caption{Decorrelation Loss}
\label{alg:uniqueloss}
\begin{algorithmic}[1]
\Require Cross-correlation matrix $C$, coefficient $\lambda$
\Function{DecorrelationLoss}{$C$, $\lambda$}
\State $l_{\text{on-diag}} = \sum_{p} (C_{pp})^2$
\State $l_{\text{off-diag}} = \sum_{p \neq q} (C_{pq})^2$
\State \Return $l_{\text{on-diag}} + \lambda \cdot l_{\text{off-diag}}$
\EndFunction
\end{algorithmic}
\end{algorithm}

\begin{algorithm}[h]
\caption{GatherDense}
\label{alg:gatherdense}
\begin{algorithmic}[1]
\Require Feature maps $\mathbf{Z}_a \in \mathbb{R}^{B \times K \times T_a}, \mathbf{Z}_b \in \mathbb{R}^{B \times K \times T_b}$, Similarity $\Delta \in \mathbb{R}^{B \times T_a \times T_b}$
\Function{GatherDense}{$\mathbf{Z}_a, \mathbf{Z}_b, \Delta$}
\State $I = \arg\max_{j} \Delta_{ij}$ 
\Comment{find best spatial correspondence}
\State $\mathbf{Z}_b \leftarrow \Call{Gather}{\mathbf{Z}_b, I}$ 
\Comment{reorder spatial features}
\State $\mathbf{Z}_a \leftarrow \Call{Reshape}{\mathbf{Z}_a, (B\cdot T_a, K)}$
\State $\mathbf{Z}_b \leftarrow \Call{Reshape}{\mathbf{Z}_b, (B\cdot T_a, K)}$
\State \Return $\mathbf{Z}_a, \mathbf{Z}_b$
\EndFunction

\end{algorithmic}
\end{algorithm}

\begin{algorithm}[h]
\caption{DenseTwins Training Procedure}
\label{alg:densetwins}
\begin{algorithmic}[1]

\Require Modality encoders $\mathcal{F}_1, \mathcal{F}_2$, projectors $\mathcal{P}_1, \mathcal{P}_2$, dataset $\mathcal{D}$, batch size $N$ 

    \For{$(\mathbf{X}_1, \mathbf{X}_2)$ in $\mathcal{D}$}

    \Statex
    \Comment{Feature extraction \& Spatial correspondences}

    \State $(\mathbf{X}_{1}', \mathbf{X}_{1}'') = \Call{Augment}{\mathbf{X}_1}$
    \State $(\mathbf{X}_{2}', \mathbf{X}_{2}'') = \Call{augment}{\mathbf{X}_2}$
    \State $\mathbf{F}_{1}', \mathbf{F}_{1}'' = \mathcal{F}_1(\mathbf{X}_{1}'), \mathcal{F}_1(\mathbf{X}_{1}'')$
    \State $\mathbf{F}_{2}', \mathbf{F}_{2}'' = \mathcal{F}_2(\mathbf{X}_{2}'), \mathcal{F}_2(\mathbf{X}_{2}'')$
    \State $(\mathbf{y}_{1}', \mathbf{Z}_{1}'), (\mathbf{y}_{1}'', \mathbf{Z}_{1}'') = \mathcal{P}_1(\mathbf{F}_{1}'), \mathcal{P}_1(\mathbf{F}_{1}'')$
    \State $(\mathbf{y}_{2}', \mathbf{Z}_{2}'), (\mathbf{y}_{2}'', \mathbf{Z}_{2}'') = \mathcal{P}_2(\mathbf{F}_{2}'), \mathcal{P}_2(\mathbf{F}_{2}'')$
    \State $\Delta_{1} = \Call{CosineSimilarity}{\mathbf{F}_{1}', \mathbf{F}_{1}''}$
    \State $\Delta_{2} = \Call{CosineSimilarity}{\mathbf{F}_{2}', \mathbf{F}_{2}''}$
    \State $\Delta_{12} = \Call{CosineSimilarity}{\mathbf{Z}_{1}'[{:}K_c], \mathbf{Z}_{2}'[{:}K_c]}$

    \Statex
    \Comment{Intra-modal global loss}

    \State $C_1^g = \frac{1}{N} \text{BN}_g(\mathbf{y}_{1}')^\top \text{BN}_g(\mathbf{y}_{1}'')$
    \State $C_2^g = \frac{1}{N} \text{BN}_g(\mathbf{y}_{2}')^\top \text{BN}_g(\mathbf{y}_{2}'')$

    \State $\mathcal{L}_g^{(1)} = \Call{BarlowTwinsLoss}{C_1^g,\lambda}$
    \State $\mathcal{L}_g^{(2)} = \Call{BarlowTwinsLoss}{C_2^g,\lambda}$

    \Statex
    \Comment{Intra-modal dense loss}

    \State $(\mathbf{Z}_{1a},\mathbf{Z}_{1b}) = \Call{GatherDense}{\mathbf{Z}_{1}',\mathbf{Z}_{1}'',\Delta_1}$
    \State $(\mathbf{Z}_{2a},\mathbf{Z}_{2b}) = \Call{GatherDense}{\mathbf{Z}_{2}',\mathbf{Z}_{2}'',\Delta_2}$

    \State $C_1^d = \frac{1}{T_1N} \text{BN}_d(\mathbf{Z}_{1a})^\top \text{BN}_d(\mathbf{Z}_{1b})$
    \State $C_2^d = \frac{1}{T_2N} \text{BN}_d(\mathbf{Z}_{2a})^\top \text{BN}_d(\mathbf{Z}_{2b})$

    \State $\mathcal{L}_d^{(1)} = \Call{BarlowTwinsLoss}{C_1^d,\lambda}$
    \State $\mathcal{L}_d^{(2)} = \Call{BarlowTwinsLoss}{C_2^d,\lambda}$

    \State $\mathcal{L}_{\text{intra}}^{(1)} = (1-\alpha)\mathcal{L}_g^{(1)} + \alpha \mathcal{L}_d^{(1)}$
    \State $\mathcal{L}_{\text{intra}}^{(2)} = (1-\alpha)\mathcal{L}_g^{(2)} + \alpha \mathcal{L}_d^{(2)}$

    \Statex
    \Comment{Cross-modal losses}

    \State $G = \frac{1}{N} \text{BN}_g(\mathbf{y}_{1}')^\top \text{BN}_g(\mathbf{y}_{2}')$

    \State $G^{ss}, G^{uu} = G[{:}K_c,{:}K_c], G[K_c{:},K_c{:}]$

    \State $L_{ss}^{g} = \Call{BarlowTwinsLoss}{G^{ss},\lambda}$
    \State $L_{uu}^{g} = \Call{DecorrelationLoss}{G^{uu},\lambda}$

    \State $(\mathbf{Z}_{1c},\mathbf{Z}_{2c}) = \Call{GatherDense}{\mathbf{Z}_{1}',\mathbf{Z}_{2}',\Delta_{12}}$

    \State $D = \frac{1}{T_1N} \text{BN}_d(\mathbf{Z}_{1c})^\top \text{BN}_d(\mathbf{Z}_{2c})$

    \State $D^{ss}, D^{uu} = D[{:}K_c,{:}K_c], D[K_c{:},K_c{:}]$

    \State $\mathcal{L}_{ss}^{d} = \Call{BarlowTwinsLoss}{D^{ss},\lambda}$
    \State $\mathcal{L}_{uu}^{d} = \Call{DecorrelationLoss}{D^{uu},\lambda}$

    \State $\mathcal{L}_{\text{cross}} = (1-\alpha)(\mathcal{L}_{ss}^{g}+\mathcal{L}_{uu}^{g})
    + \alpha(\mathcal{L}_{ss}^{d}+\mathcal{L}_{uu}^{d})$

    \Statex
    \Comment{Total loss}

    \State $\mathcal{L} = (\mathcal{L}_1 + \mathcal{L}_2 + \mathcal{L}_{\text{cross}})/3$
    \State $\Call{Backpropagation}{\mathcal{L}}$

\EndFor

\end{algorithmic}
\end{algorithm}

\section{Analysis of the Learned Decomposition}
\label{sec:supp-analysis}

While downstream performance provides the primary evaluation of pretrained representations, it does not directly verify whether the proposed decomposition successfully separates shared and modality-specific information. To better understand the behaviour of DenseTwins, we analyze the learned embedding space through three complementary perspectives: correlation structure, cross-modal retrieval, and feature visualization.

\subsection{Correlation Structure}

DenseTwins explicitly encourages shared dimensions to capture information that is consistent across modalities while suppressing correlations between modality-specific components. To verify that this behaviour emerges in practice, we analyze the global cross-correlation matrix $G$ computed between modalities after pretraining.

In particular, we compare the diagonal values of the shared correlation block $G^{ss}$ and the modality-specific block $G^{uu}$. We expect dimensions belonging to the shared subspace to exhibit strong cross-modal correlations, whereas dimensions belonging to the modality-specific subspace should remain largely decorrelated.

The observed behaviour is consistent with the objective. The average diagonal value of $G^{ss}$ is $0.382$, while the average diagonal value of $G^{uu}$ is only $0.038$, indicating that most cross-modal information is concentrated in the shared subspace and that modality-specific dimensions remain largely independent across modalities.

\subsection{Cross-Modal Retrieval}

We further evaluate whether the shared subspace captures semantically meaningful cross-modal information through a patch-level retrieval experiment.

Given a query RGB patch embedding, we retrieve the most similar Depth patch embedding using either the shared or the modality-specific subspace. A retrieval is considered correct if the retrieved patch corresponds to the same scene and spatial location.

Results are reported in Table~\ref{tab:retrieval}. Retrieval using the shared subspace substantially outperforms retrieval using the modality-specific subspace, achieving Recall@1 of $0.62$ and Recall@5 of $0.85$. In contrast, retrieval based on the modality-specific dimensions performs close to chance level. These results provide direct evidence that semantic correspondences between modalities are encoded primarily in the shared subspace.

\begin{table}[h]
\centering
\caption{Cross-modal patch retrieval using different subspaces.}
\label{tab:retrieval}
\begin{tabular}{lcc}
\toprule
Subspace & Recall@1 & Recall@5 \\
\midrule
Shared & \textbf{0.62} & \textbf{0.85} \\
Specific & 0.06 & 0.13 \\
\bottomrule
\end{tabular}
\end{table}

\subsection{Feature Visualization}

\begin{figure}[h]
\centering
\includegraphics[width=0.7\linewidth]{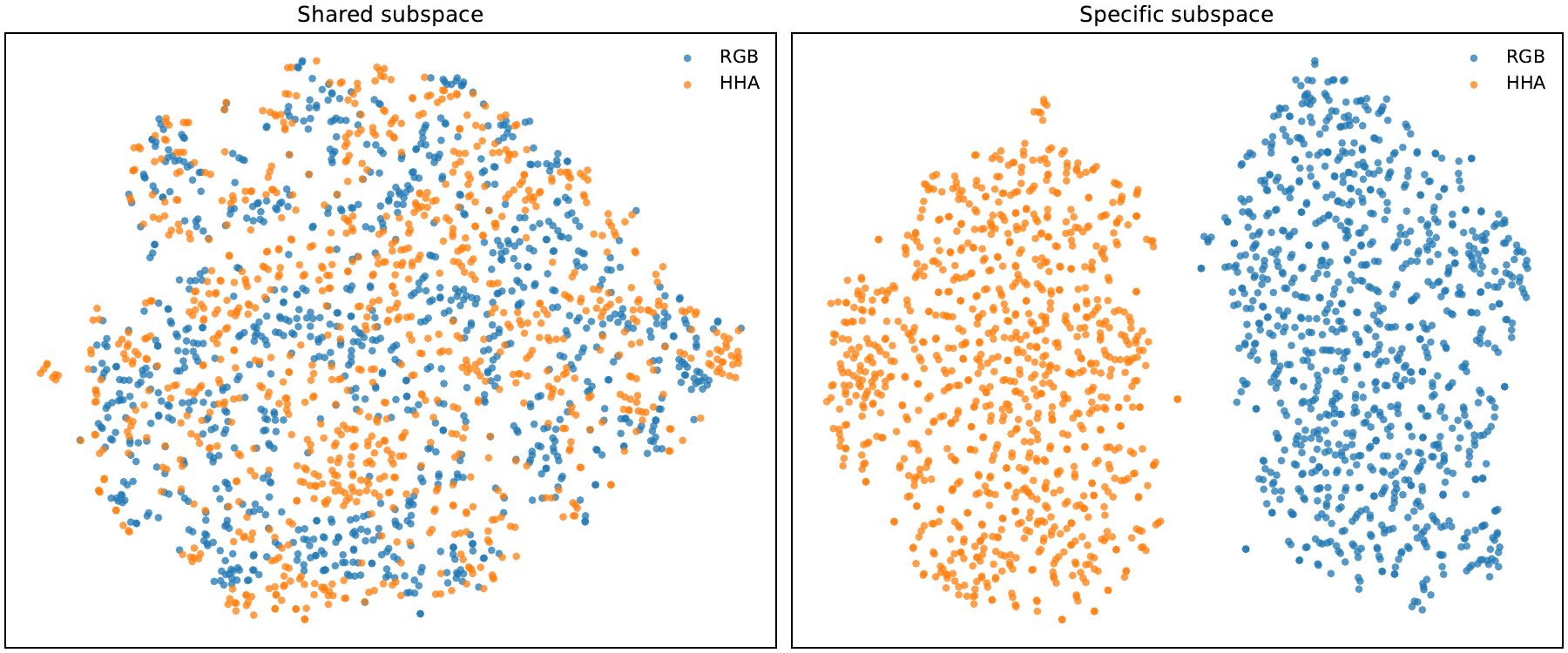}
\caption{t-SNE visualization of global embeddings extracted from the shared and modality-specific subspaces. The shared subspace exhibits stronger cross-modal alignment, whereas the modality-specific subspace remains separated according to modality.}
\label{fig:subspace}
\end{figure}

Finally, Figure~\ref{fig:subspace} presents t-SNE projections of the global embeddings extracted from the shared and modality-specific subspaces.

Embeddings from the shared subspace exhibit substantially stronger cross-modal mixing, indicating that samples from different modalities are projected to a common semantic space. In contrast, embeddings from the modality-specific subspace remain clearly separated according to their sensing modality, suggesting that modality-dependent information is preserved in these dimensions.

Together, the correlation analysis, retrieval experiment, and feature visualization consistently support the intended behaviour of DenseTwins and provide evidence that the learned embedding space successfully disentangles shared and modality-specific information. Combined with the consistent downstream improvements reported in Section~\ref{sec:experiments}, these results suggest that dense modality decoupling is not only achieved in practice, but also contributes to learning more effective multimodal representations for dense prediction tasks.

\vspace{-3mm}

\section{Additional Analysis}
\label{sec:supp-additional-analysis}

\subsection{Computational Overhead}
\label{sec:supp-computational-overhead}

We analyze the computational and memory overhead introduced by the projection heads and the dense matching mechanism during pretraining, to complement the ablation studies presented in \Cref{sec:experiments} with a direct measurement of their training cost.

We profile a full training step of DenseTwins on a single NVIDIA A100 GPU, decomposing the forward and backward computation into its main stages for two per-GPU batch sizes (8 and 24). \Cref{tab:compute-breakdown} reports, for each stage, its wall-clock cost in milliseconds together with its share of the total forward pass and of the full training step (forward, backward and optimizer update).

\begin{table}[t]
\centering
\small
\setlength{\tabcolsep}{3pt}
\caption{Per-step time breakdown of a DenseTwins training step (single A100 GPU) at two per-GPU batch sizes.}
\label{tab:compute-breakdown}
\begin{tabular}{lrrr}
\toprule
Stage & ms & \% forward & \% step \\
\midrule
\multicolumn{4}{l}{\textit{Batch size 8/GPU}} \\
\midrule
Backbones forward & 69.9 & 56.8 & 17.7 \\
Projection heads & 9.3 & 7.6 & 2.4 \\
Dense Correspondences & 3.6 & 2.9 & 0.9 \\
BT correlation and loss & 41.0 & 33.4 & 10.4 \\
Total forward & 123.0 & 100.0 & 31.2 \\
Backward step & 270.7 & -- & 68.8 \\
Total & 394.0 & -- & 100.0 \\
\midrule
\multicolumn{4}{l}{\textit{Batch size 24/GPU}} \\
\midrule
Backbones forward & 200.4 & 65.3 & 21.2 \\
Projection heads & 21.9 & 7.1 & 2.3 \\
Dense Correspondences & 8.9 & 2.9 & 0.9 \\
BT correlation and loss & 76.5 & 24.9 & 8.1 \\
Total forward & 307.0 & 100.0 & 32.5 \\
Backward step & 639.0 & -- & 67.6 \\
Total & 946.0 & -- & 100.0 \\
\bottomrule
\end{tabular}
\end{table}

The backbones' forward pass, executed four times per step (one pass per augmented view and per modality), together with the backward step, dominates the training step at both batch sizes. In contrast, the components introduced by DenseTwins on top of a standard multimodal pretraining pipeline add only a modest overhead to the forward pass: the projection heads account for $7.1$--$7.6\%$ of the forward pass ($2.3$--$2.4\%$ of the full step, not counting the backward pass through the projectors themselves), while the dense correspondences step, which establishes the spatial matching used by the dense objectives, accounts for only $2.9\%$ of the forward pass ($0.9\%$ of the full step).

Considering the full training step, including the backward pass through the projection heads, the projection heads increase training time by $+25\%$ and peak GPU memory by $+26\%$ with respect to a version of the pipeline without them. \Cref{tab:projector-overhead} quantifies the practical impact of this overhead on a representative large-scale pretraining run on DSEC using $8\times$A100 GPUs.

\begin{table}[t]
\centering
\small
\caption{Effect of the projection heads on DSEC pretraining using $8\times$A100 GPUs. At a fixed memory budget, removing them allows a larger per-GPU batch size and reduces training time.}
\label{tab:projector-overhead}
\begin{tabular}{lcc}
\toprule
Configuration & Batch size/GPU & Training time \\
\midrule
w/o projection heads & 28 & 5.1\,h \\
w/ projection heads & 22 & 6.4\,h \\
\bottomrule
\end{tabular}
\end{table}

We stress that this overhead is incurred only during pretraining: the projection heads are always discarded before transferring the encoders to downstream tasks (\Cref{sec:implementation-details-supp}), so downstream fine-tuning/probing cost and inference cost of the encoders are unaffected by it. Removing the projection heads, as in the \textit{w/o projection heads} configuration of \Cref{tab:projector-overhead}, means that shared/specific decoupling and dense correspondence matching are instead computed directly on the backbone feature space rather than on a dedicated projected embedding. As shown in the ablation studies in \Cref{sec:experiments} (\Cref{tab:ablation-projector,tab:ablation-losses}), using dedicated projection heads is nonetheless beneficial for the performance of DenseTwins, motivating their use despite the added training-time cost.

\subsection{Single Modal Results}
\label{sec:supp-missing-modalities}

DenseTwins is a multimodal method: both modalities are required during pretraining, as the shared/specific decomposition is learned jointly across the modality pair. However, DenseTwins learns two separate modality-specific encoders rather than a single fused backbone, which leaves room for flexible deployment solutions, including deploying a single-modal model built from one of the two pretrained encoders when only one modality is available at test time. To assess how competitive the resulting single-modal representations are, we take the RGB and Depth encoders jointly pretrained by DenseTwins on SUN-RGBD and evaluate each of them individually, and compare them against dedicated single-modal SSL baselines (DINO, iBOT, I-JEPA, LeJEPA) trained separately on each modality with the same backbone. We also report the multimodal setting, where the two modality-specific encoders are combined via late fusion (feature concatenation), for both DenseTwins and the baselines. All representations are evaluated with linear probing (LP) and a lightweight FCN head on frozen features, following the protocol of \Cref{sec:experiments}.

\Cref{tab:missing-modality} reports the results. When only a single modality is available, the DenseTwins encoder pretrained for that modality remains competitive with dedicated single-modal baselines: it is the best-performing method on Depth LP, and consistently outperforms DINO, iBOT and LeJEPA on RGB (both LP and FCN), although I-JEPA attains higher RGB scores and DINO attains a slightly higher Depth FCN score. Notably, in the multimodal (Concat) setting, DenseTwins outperforms all baselines, including I-JEPA, on both LP and FCN, despite I-JEPA's RGB-only encoder being stronger than the corresponding DenseTwins encoder. This suggests that the DenseTwins representation captures modality-specific information more effectively than single-modal baselines once combined across modalities, even where its individual single-modal encoders are not the strongest in isolation.

\begin{table}[t]
\centering
\small
\setlength{\tabcolsep}{3pt}
\caption{Linear probing (LP) and FCN performance (mIoU) on SUN-RGBD when only RGB or only Depth is available at test time, compared to the multimodal (Concat) setting obtained via late fusion.}
\label{tab:missing-modality}
\begin{tabular}{lcccccc}
\toprule
& \multicolumn{2}{c}{RGB} & \multicolumn{2}{c}{Depth} & \multicolumn{2}{c}{Concat} \\
\cmidrule(lr){2-3}\cmidrule(lr){4-5}\cmidrule(lr){6-7}
Method & LP & FCN & LP & FCN & LP & FCN \\
\midrule
DenseTwins & 17.2 & 26.7 & \textbf{27.8} & 32.5 & \textbf{33.7} & \textbf{40.6} \\
I-JEPA & \textbf{19.6} & \textbf{27.5} & 27.2 & 32.0 & 31.1 & 35.2 \\
DINO & 15.0 & 22.8 & 26.4 & \textbf{33.9} & 31.0 & 37.6 \\
iBOT & 12.7 & 19.7 & 24.8 & 31.6 & 30.8 & 36.3 \\
LeJEPA & 9.5 & 19.1 & 17.6 & 30.0 & 23.9 & 32.7 \\
\bottomrule
\end{tabular}
\end{table}

\subsection{Dense Matching Robustness}
\label{sec:supp-dense-matching-robustness}

The retrieval experiment in \Cref{sec:supp-analysis} (\Cref{tab:retrieval}) measures cross-modal correspondence quality under perfectly aligned modalities. To assess robustness to imperfect spatial alignment between modalities, we repeat the same retrieval protocol after applying synthetic misalignment to one modality: a random translation of one patch (16 pixels) or two patches (32 pixels), before performing retrieval.

\Cref{tab:retrieval-misalignment} reports the results. Recall@1 and Recall@5 remain close to the aligned setting under both shifts ($0.60$--$0.61$ and $0.82$--$0.84$, respectively, vs.\ $0.62$/$0.85$ without misalignment), showing that the dense correspondence mechanism degrades only marginally under moderate spatial misalignment between modalities.

\begin{table}[h]
\centering
\small
\caption{Cross-modal patch retrieval recall on SUN-RGBD under synthetic misalignment (translation applied to one modality) using the shared subspace.}
\label{tab:retrieval-misalignment}
\begin{tabular}{lcc}
\toprule
Misalignment & Recall@1 & Recall@5 \\
\midrule
None (aligned) & 0.62 & 0.85 \\
Shift 16px (1 patch) & 0.60 & 0.84 \\
Shift 32px (2 patches) & 0.61 & 0.82 \\
\bottomrule
\end{tabular}
\end{table}

This robustness follows directly from a design choice of DenseTwins: cross-modal correspondences are established through a nearest-neighbour search in feature space over \emph{all} spatial positions, rather than by assuming a fixed positional correspondence between the two modalities. It is precisely this relaxation, matching by feature similarity rather than by position, that makes the mechanism robust to the kind of spatial misalignment a strictly positional matching scheme would be sensitive to.

\section{Qualitative Downstream Results}
\label{sec:supp-qualitative}

\begin{figure*}[t]
\centering
\includegraphics[width=\textwidth]{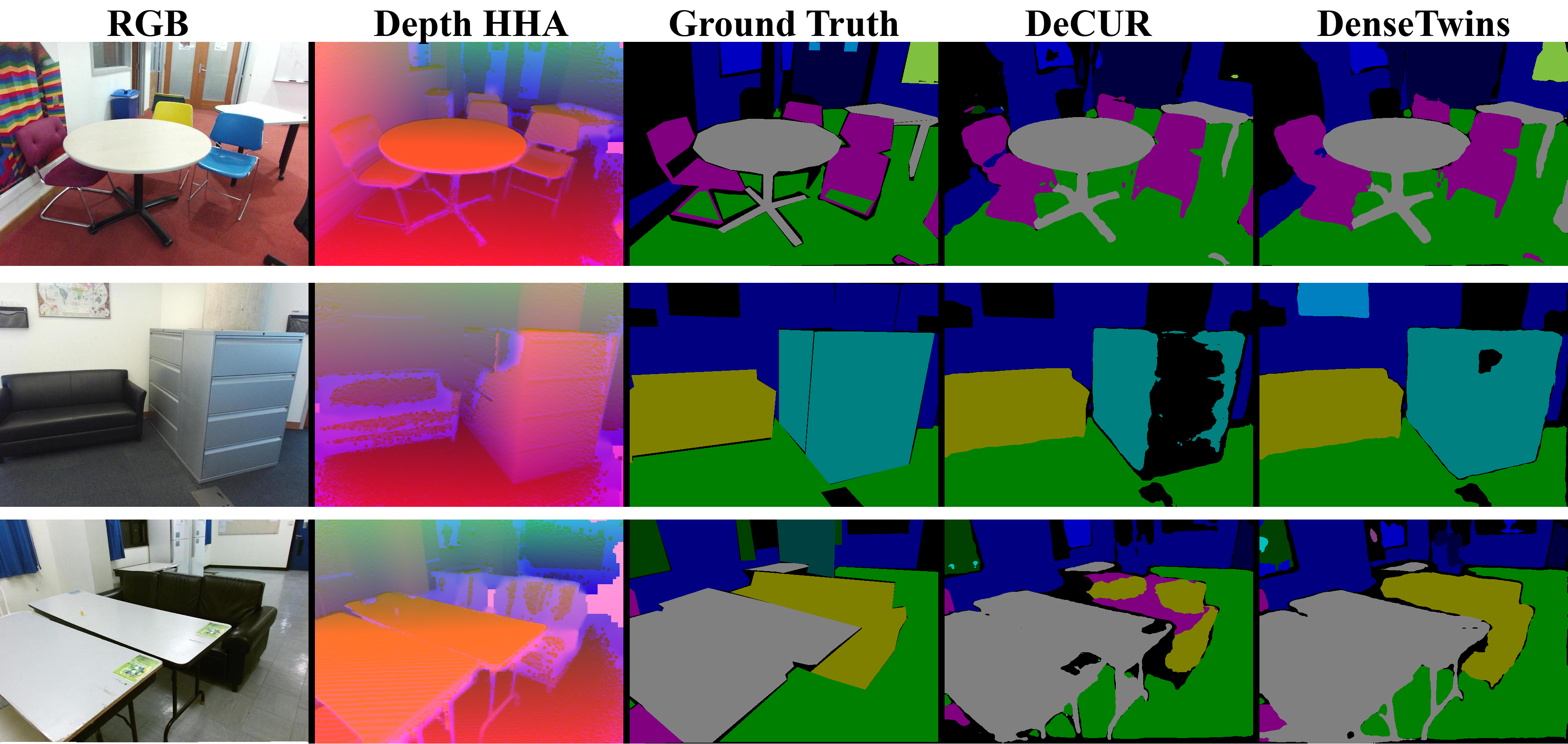}
\caption{\textbf{Qualitative Results on SUN-RGBD}. From left to right we show the RGB image, the depth modality encoded as HHA, the ground-truth semantic segmentation, and the predictions obtained using DeCUR and DenseTwins pretraining.
While both methods correctly segment the main scene structures, DenseTwins produces sharper object boundaries and more consistent predictions.}
\label{fig:sunrgbd}
\end{figure*}

\begin{figure*}[t]
\centering
\includegraphics[width=\textwidth]{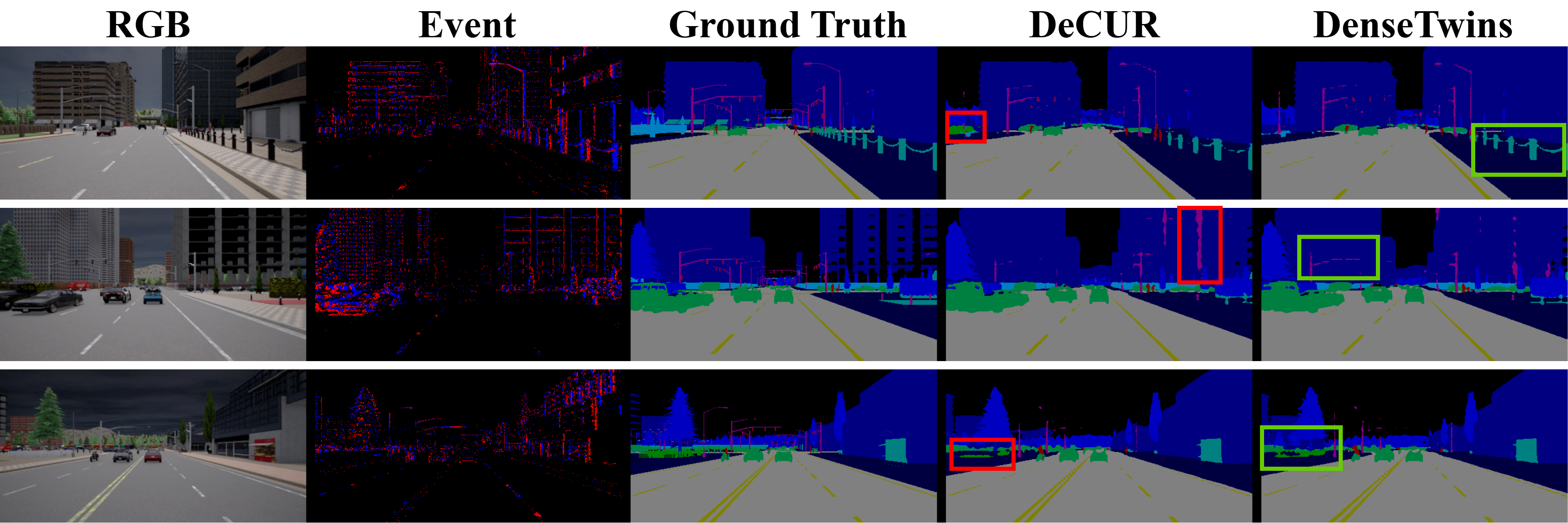}
\caption{\textbf{Qualitative Results on Eventscape}. From left to right, we show the RGB frame, the event representation, the ground-truth segmentation, and the predictions obtained using DeCUR and DenseTwins. Highlighted regions show typical failure cases of DeCUR (red boxes), such as missed distant objects or fragmented thin structures, while DenseTwins provides more complete predictions and improved detection of small or distant regions (green boxes).}
\label{fig:eventscape}
\end{figure*}

We present qualitative results of the downstream semantic segmentation models on two representative multimodal datasets: SUN-RGBD and EventScape.
For each example, we visualize the input modalities, the ground-truth segmentation, and the predictions obtained using backbones pretrained with DeCUR and with the proposed DenseTwins framework.

\textbf{SUN-RGBD.}
\Cref{fig:sunrgbd} shows qualitative segmentation results on indoor RGB-D scenes.
While both methods correctly segment the main structures of the scene, DenseTwins generally produces cleaner object boundaries and more consistent predictions across objects with similar appearance.
In particular, DenseTwins better preserves thin structures and object contours (e.g., chairs and table legs) and reduces spurious predictions around object boundaries.
These results suggest that learning dense multimodal representations improves the spatial consistency of the learned features.

\textbf{EventScape.}
\Cref{fig:eventscape} shows qualitative results on driving scenes combining RGB images and event representations. 
For visualization purposes, the event modality is displayed using a frame-based representation where positive and negative events are accumulated into separate channels. 
However, the model is trained using a voxel-based event representation as input, as described in \Cref{sec:implementation-details-supp}. 
In these scenarios, DenseTwins better captures distant objects and thin structures such as poles and fences, which are often missed or fragmented by the baseline, improving the recognition of small objects in the scene.
These results suggest that dense multimodal alignment enables the model to better exploit complementary cues from RGB and event data, particularly for dynamic and fine-grained structures.

\end{document}